\documentclass[10pt,journal,compsoc]{IEEEtran}
\usepackage{dirtytalk}
\usepackage{times}
\usepackage{epsfig}
\usepackage{graphicx}
\usepackage{amsmath}
\usepackage{amssymb}
\usepackage{color}
\usepackage{xspace}
\usepackage{subfigure}

\ifCLASSOPTIONcompsoc
  \usepackage[nocompress]{cite}
\else
  \usepackage{cite}
\fi
\ifCLASSINFOpdf
\else
\fi
\begin{document}
%
\title{DeepCloth: Neural Garment Representation for Shape and Style Editing}
%
%
%
%

\author{Zhaoqi~Su,
        Tao~Yu,
        Yangang~Wang,~\IEEEmembership{Member,~IEEE,}
        and~Yebin~Liu,~\IEEEmembership{Member,~IEEE}
\IEEEcompsocitemizethanks{\IEEEcompsocthanksitem Z. Su, T. Yu and Y. Liu are with Tsinghua University, Beijing, China.\protect\\
\IEEEcompsocthanksitem Y. Wang is with Southeast University, Nanjing, China.\protect\\
\IEEEcompsocthanksitem Corresponding authors: Yebin Liu}
\thanks{This work was supported by NSFC under Grants 62125107 and 62171255, in part by the National Key R\&D Program of China under Grant 2021ZD0113503 and in part by China Postdoctoral Science Foundation under Grant 2020M670340.}
\thanks{Digital Object Identifier no. 10.1109/TPAMI.2022.3168569}
}

\newcommand{\ourpaper}{DeepCloth\xspace}
\newcommand{\ParamNet}{ParamNet\xspace}
\newcommand{\AnimNet}{AnimNet\xspace}
\newcommand{\InferNet}{3DInferNet\xspace}
\newcommand{\etal}{et al.\xspace}

\IEEEtitleabstractindextext{%
\begin{abstract}
   Garment representation, editing and animation are challenging topics in the area of computer vision and graphics. 
   It remains difficult for existing garment representations to achieve smooth and plausible transitions between different shapes and topologies. 
   In this work, we introduce, \ourpaper, a unified framework for garment representation, reconstruction, animation and editing.
   Our unified framework contains 3 components:
   First, we represent the garment geometry with a \say{topology-aware UV-position map}, which allows for the unified description of various garments with different shapes and topologies by introducing an additional topology-aware UV-mask for the UV-position map. 
   Second, to further enable garment reconstruction and editing, we contribute a method to embed the UV-based representations into a continuous feature space, which enables garment shape reconstruction and editing by optimization and control in the latent space, respectively. 
   Finally, we propose a garment animation method by unifying our neural garment representation with body shape and pose, which achieves plausible garment animation results leveraging the dynamic information encoded by our shape and style representation, even under drastic garment editing operations. 
   To conclude, with \ourpaper, we move a step forward in establishing a more flexible and general 3D garment digitization framework. 
   Experiments demonstrate that our method can achieve state-of-the-art garment representation performance compared with previous methods. 
\end{abstract}

}

\maketitle

\IEEEraisesectionheading{\section{Introduction}\label{sec:intro}}

\IEEEPARstart{3}{D} garment representation, modeling, editing and animation/simulation have numerous applications in clothing design, digital humans, and virtual try-on. 
Traditional high-fidelity 3D garment modeling and animation often rely on artist design or heavy simulation methods, such as physically based simulation~\cite{PBSMS95_provot1995deformation}, which consume enormous labor costs or computational resources. 
In recent years, neural garment representations based on deep learning techniques have achieved impressive garment modeling or animation results~\cite{DeepWrinkles_Lahner_2018_ECCV, GarNet++_Gundogdu_TPAMI, CAPE_Ma_2020_CVPR, PixelBased_Jin_SCA_2020, SIZER_tiwari20sizer, SewingPatterns_Yu_2020_ECCV}. However, the majority of these methods either focus on encoding garment dynamics for specific clothing (clothing-specific learning) or aim at 3D clothing recovery from images without any editing capacities. 
This is because establishing a unified framework for shape/style editable garment representation, reconstruction and animation remains challenging.
Although the most recent neural garment modeling/animation work TailorNet~\cite{TailorNet_Patel_2020_CVPR} achieves impressive detailed clothing dynamics recovery for different human shapes and poses, it still defines garments on top of a predefined fixed template, in which the garment topology is fixed. Such a fixed representation 
limits its ability to achieve an ideal garment editing framework, e.g., enabling transition from long pants to shorts or from front-opening T-shirts to front-closing shirts. 
Concurrent work~\cite{SIMPLicit_Corona_2021_CVPR} by Corona \etal focuses more on establishing garment shape/style representations and less on shape-/style-dependent garment animation based on their styles, 
while our method further proposes a garment shape-dependent animation module, which shows more dynamics while performing animation with different garment styles.

\begin{figure}[t]
    \centering
    \centering
    \includegraphics[width=0.85\linewidth]{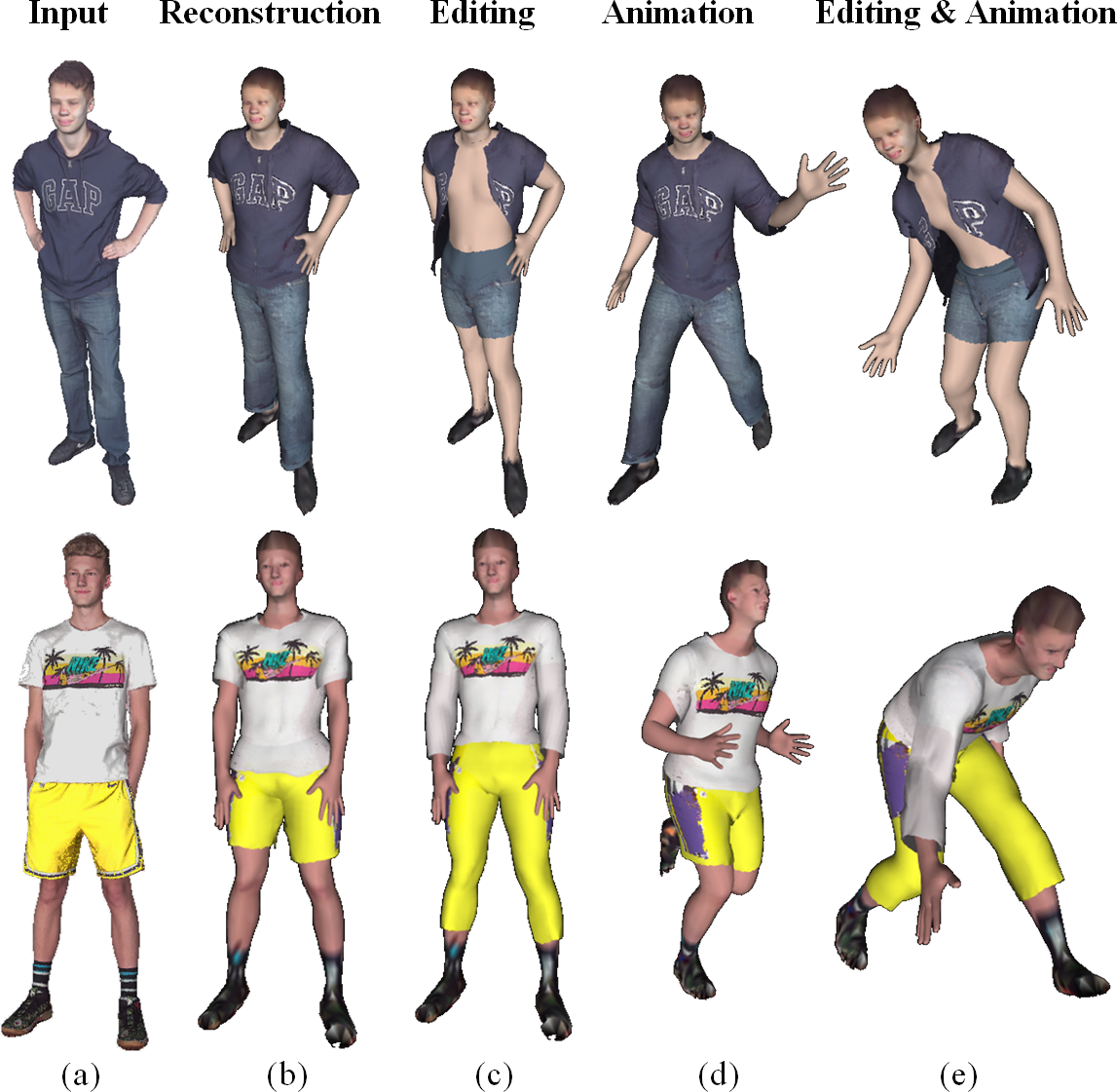}
    \caption{
    Our neural garment representation framework, DeepCloth, enables garment reconstruction (b), shape editing (c) (from close to open, from short to long, from loose to tight, etc.) and animation (d,e) (with garment-specific dynamics even after significant topological editing of the garment) given a 3D scan with arbitrary pose (a). 
    }
    \label{fig:teaser}
\end{figure}

In this paper, we argue that it is essential to learn a compact and uniform space for garments with different shapes and styles,
which will form a unified garment representation framework, and then be further used for garment reconstruction and animation. Such a representation should enable free and smooth style transitions between different garment shapes and styles, even for garments with different topologies, e.g., from front-opening clothes to front-closing clothes.
By transferring the garment representation into a neural feature space, and mapping the 3D scanned garment mesh into the same feature space, such a representation can also be used to perform garment animation and 3D garment shape editing using deep neural networks as demonstrated in Fig.~\ref{fig:teaser}. However, fulfilling such a representation is challenging due to the large topological changes and nonuniform latent space encoding.


In this paper, we propose \ourpaper, a unified framework for garment representation, reconstruction, animation and editing by assembling different garment shapes and topologies into a unified representation framework.
Technically, we propose a \say{topology-aware UV-position map} representation to encode both the topologies and the geometric details of garments. To achieve plausible garment transitions under different topologies, we transform the UV binary mask into a continuous distance map. By encoding the UV-position map with a transformed mask into a feature space using our proposed \ParamNet, we can represent garments with different shape styles and topologies in a unified network and thus achieve flexible garment editing and continuous shape and style transitions between different garments by feature interpolation and decoding. 
As the \say{UV map} is widely used to represent 3D shapes (such as Tex2shape~\cite{Tex2Shape_Alldieck_2019_ICCV} for human modeling), we believe that our introduction of the \say{topology-aware-mask} for UV-map-representation, as well as the demonstration of neural shape editing capacity, may inspire future 3D mesh/shape/texture editing studies related to the use of UV maps.

With our deep learning network trained on the large-scale synthetic dataset of 3D clothed human sequences with various garment styles, i.e. CLOTH3D~\cite{CLOTH3D_2020_ECCV}, we can parameterize clothing shape variations of front-opening T-shirts, T-shirts, shirts, pants, skirts, dresses, and jumpsuits. Additionally, with our proposed animation module \AnimNet and 3D-shape inference module \InferNet, \ourpaper can generate 4D sequences of garment dynamics (see Fig.~\ref{fig:teaser}(d,e)) or extract the clothing shape parameters from a clothed human model under arbitrary poses (see Fig.~\ref{fig:teaser}(b)), which enhances its ability for 3D clothing shape editing (see Fig.~\ref{fig:teaser}(c)). The main contributions of this work are summarized as follows:

\begin{itemize}
    \item We propose a unified garment modeling framework based on a UV-mask garment representation, which further enables garment reconstruction, animation and editing.
    
    \item We propose a topology-aware continuous UV-mask neural garment representation and encode such representation into a unified continuous feature space, which enables joint learning of both the 3D position and the topology of the garments, and neural control of garment shape and topologies. (Sect.~\ref{subsec:garment_repre},~\ref{subsec:garment_param})
        
    \item By mapping the garment 3D information onto the garment feature space, or unifying the proposed garment representation with human shape and pose information, we can perform garment shape reconstruction and animation based on our neural garment representation, which can generate plausible garment dynamics even under drastic garment editing operations. (Sect.~\ref{subsec:garment_infer},~\ref{subsec:garment_anim})

\end{itemize}
\section{Related Work}
\label{sec:related}

There are numerous works on garment representation, animation and reconstruction. Here, we mainly review the works that are most related to our approach.

\textbf{Garment representation and animation.} There are essentially three approaches for garment animation: physics-based simulation (PBS), data-driven methods, and animation based on capture. 

For physics-based simulation (PBS), traditional physics based garment simulation formulates the garment as a mass-spring system with force-based simulation~\cite{PBSMS95_provot1995deformation, PBSMS05_choi2005stable, PBSMS13_liu2013fast} or other physical models based on the finite element method~\cite{PBSFE97_pbsbonet1997nonlinearcontinuummechanics, PBSFE17_Jiang:2017:Anisotropic}, with the explicit Euler method~\cite{PBSEUE07_press2007numerical} or implicit/semi-implicit Euler method~\cite{PBSEUI87_terzopoulos1987elastically, PBSEUI98_baraff1998large, PBSEUE07_press2007numerical}. These methods can generate realistic clothing with vivid dynamics given a designed garment shape and garment template, but mostly incur considerable computational costs for numerous integration iterations for clothing dynamics, and cannot perform a more general shape control of the garment.

Data-driven methods aim to shorten the computational time for garment animation with a more flexible garment representation. Early methods such as ~\cite{DD2010_examplebased, DD2010_stablespace, DD2013_nearexhaustive} use a nearest neighbor search or linear regression to animate clothing on the human body with different poses and shapes. Recent works have mainly adopted deep learning methods to perform garment animation.~\cite{Sharespace_garmentdesign_Wang_SA18} learns a shared space for garment style variation, and can predict garment shape from a user sketch with a fixed pose.~\cite{virtualtryon_Santesteban19, AnaCloDeform_Yang_2018_ECCV} regress the garment shape with various human poses and shapes with MLP or RNN methods.~\cite{GarNet_Gundogdu_2019_ICCV, GarNet++_Gundogdu_TPAMI, CAPE_Ma_2020_CVPR} propose garment animation by 3D garment draping or SMPL-based garment deformation, using a graph convolution network to obtain garment shapes worn on a human model with different shapes and poses.~\cite{DeepWrinkles_Lahner_2018_ECCV, PixelBased_Jin_SCA_2020} propose pixel-based garment 2D representation based on texture mapping on a human model or on a template-based texture space, which is similar to our representation method, but they cannot generate the shape parameters of the garments.~\cite{360DegreeTexture_lazova2019360} leverages the human parsing of the image to mask the UV-map representations of garments and controls the garment shape by editing the masks. However, without compact encoding of the garment representation UV map and masks, it barely performs continuous garment style transition, and the garment animation can only be performed through skinning, without leveraging the masks to infer shape dynamics. Additionally, it cannot represent garments that are not homotopy to human models such as dresses.~\cite{SewingPatterns_Yu_2020_ECCV} can interpolate between different garment styles, but it can only interpolate the \say{sewing patterns}, meaning that ~\cite{SewingPatterns_Yu_2020_ECCV} only interpolates the area of the body covered by the garments, without a general shape parametrization framework for generating more garment shapes. These methods use garment vertices, graph-based garment representation or pixel-based garment representation to perform garment animation but lack garment shape parametrization modules for flexibly and conveniently controlling the garment shape. In addition, in regard to computer vision tasks such as relighting on clothed humans,~\cite{NLT_10.1145/3446328} proposes a novel approach for representing the lighting environment and view visibility in UV space, which leverages the flexibility of UV representations for performing realistic and high-quality relighting and view synthesis on real captured data of humans.

\begin{figure*}[t]
    \centering
    \includegraphics[width=0.98\textwidth]{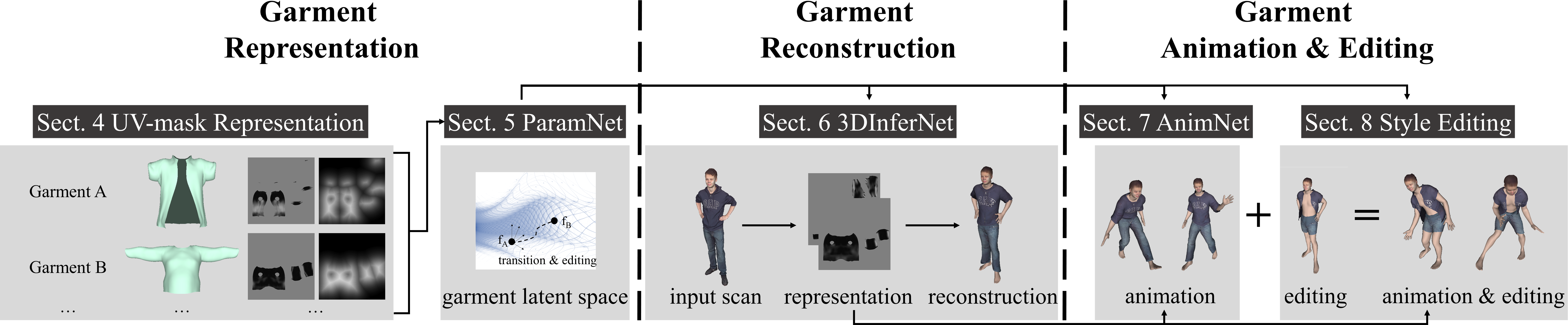}
    \caption{The demonstration of our \ourpaper framework. From left to right: garment representation, garment reconstruction and garment animation module.}
    \label{fig:framework}
\end{figure*}

For garment animation based on real capture, this kind of method for garment animation focuses on recovering the static or dynamic garment shape from a given picture or video.~\cite{CAPMV08_Bradley:MarkerlessGarmentCap, CAPMV09_Popa:2009, CAPMV10_AnimatableHuman} recover garment shapes from multiview stereo,~\cite{ClothCap_Pons-Moll:2017:ClothCap, DeepWrinkles_Lahner_2018_ECCV} recover garment dynamics from 4D sequences,~\cite{GarmentModeling15_Chen:2015:GMD:2816795.2818059, SimulCap_Yu_2019_CVPR} propose a system for garment shape recovery from a single RGBD camera, while~\cite{MulayCap_Su_2020_TVCG} proposes a method for extracting the garment template shape and recovering garment dynamics from a single RGB input. Recently,~\cite{MGN_Bhatnagar_2019_ICCV} propose a method for extracting multiple garments from several input images and dressing them on other human bodies, while~\cite{HMR_Kanazawa_2018_CVPR, HMR+_Alldieck_2019_CVPR} propose a CNN-based method for recovering the human shape and pose. Such methods express the garment as a deformation of the subset of the human model, and it is difficult to express more types of garments like dresses and loosened front-opening clothing. Alldieck \etal~\cite{Tex2Shape_Alldieck_2019_ICCV} proposes a 2D texture-based human with a garment shape representation method for recovering the whole shape from a single image.~\cite{LiveCap_10.1145/3311970, DeepCap_Habermann_2020_CVPR} propose monocular human performance capture methods that can recover human motion and garment geometry details, given monocular RGB video inputs. These methods mainly focus on recovering garment shapes from input images, without proposing a general garment shape representation framework.

\textbf{Garment shape parametrization.} Recently, a few works focus on establishing a garment shape parametrization framework. Shen \etal~\cite{SewingPatterns_Yu_2020_ECCV} demonstrates the garment style interpolation results, but it only controls the change of the covering area on the human body, without generating a general shape expression. Tiwari \etal~\cite{SIZER_tiwari20sizer} proposes a framework for parsing the 3D input to extract the garment shape and change the size of the garment, but in view of shape parametrization, it only controls one dimension of the garment shape. TailorNet~\cite{TailorNet_Patel_2020_CVPR} proposes a garment shape parametrization and animation framework; however, as mentioned in Sect.~\ref{sec:intro}, with the \say{offset on template vertices} expression, it is difficult for TailorNet~\cite{TailorNet_Patel_2020_CVPR} to generalize to more types of garment topology, such as front-opening garments and long dresses. In addition, it shows limited ability to perform large garment shape changes, e.g., from long trousers to shorts or from long dresses to skirts. Additionally, compared to our \ourpaper, it has less capability for performing 3D garment shape inference and flexible 3D shape editing. Therefore, it does not meet the demand for establishing a general framework for garment representation enabling garment shape and style transition. Meanwhile, our \ourpaper proposes a general garment shape representation framework, which enables more general 3D garment reconstruction, animation and editing.
\section{Overview}
\label{sec:overview}

Our goal is to establish a unified framework for garment representation, reconstruction, animation and editing, as shown in Fig.~\ref{fig:framework}. 
We first introduce the idea behind designing the whole framework. For a unified garment modeling model, traditional methods often rely on a fixed garment mesh template, which can hardly be applied to a general garment style and shape representation. To allow for a flexible and neural editable garment modeling framework, we propose our UV-mask garment representation.
Together with our proposed CNN-based \ParamNet, a unified and compact garment style space is established. Furthermore, to perform garment shape inference, animation and shape editing supporting various styles of garments, we propose different CNN- and PointNet-based networks, which establish mappings between different data domains, e.g., T-posed and animated garment shapes, or 3D mesh space and 2D garment UV space. 

Methodically, 
we first propose a UV-position map with continuous mask representation, in which the mask denote the topology and covering areas of the garments, while the rendered texels on the UV map denotes the geometry details (see Sect.~\ref{subsec:garment_repre}). Such a representation transfers the garment shape style and topology into a 2D UV-map, which is naturally suitable for the continuous transition between different garment shapes. Then, we perform UV map encoding by introducing \ParamNet, which maps both the UV map and its mask information into a feature space by using a CNN-based encoder-decoder structure (see Sect.~\ref{subsec:garment_param}). By changing and interpolating the features in the feature space, garment shape transition and editing can be performed and can be applied to the following garment shape inference and animation module (see Sect.~\ref{subsec:garment_infer} and Sect.~\ref{subsec:garment_anim}). 
Specifically, given 3D garment scans, the garment inference module can reconstruct the garment shape by transforming the point clouds to our neural garment representation (Sect.~\ref{subsec:garment_infer}). With the garment shapes mapped onto the garment feature space, garment animation or shape editing can be applied to the reconstructed garments. (Sect.~\ref{subsec:garment_anim},~\ref{subsec:garment_edit}). Note that our garment animation module is able to learn specific garment dynamic information according to different garment topologies, which leads to more plausible 4D garment animation results, even under drastic garment editing operations.

\section{T-posed Garment Representation}
\label{subsec:garment_repre}

For T-posed garment representation, a continuous UV-mask garment representation is proposed to map the 3D garment mesh onto the continuous 2D UV space, as illustrated below. Such a representation naturally encodes the 3D garment geometry distribution on top of human bodies, without relying on a fixed template, therefore supporting style transition and editing among different garment topologies. 

The first step of our \ourpaper is to represent garments with different shapes and topologies in a compact space. Therefore, different garment shape styles of a garment type, i.e., front-opening/front-closing T-shirts with long/short sleeves, can be mapped into the same feature space. Note that this section will only deal with the T-posed garment model for garment shape encoding and transition, while garments under arbitrary poses will be handled in Sect.~\ref{subsec:garment_anim}. 

In our representation framework, we regard a garment mesh as a geometric structure covering the human surface and then map such clothing to a standard human model UV map~\cite{VideoBased3DPeople_Alldieck_2018_CVPR} which stores the garment topologies and normal distance from the human body. By mapping the 3D garment geometry onto the 2D SMPL UV space, we are able to establish the relationship between the garment vertices and the nearby vertices on the human model, and better represent the geometric features.


    

\begin{figure}[ht]
    \centering
    \subfigure[]{
    \begin{minipage}[t]{0.47\linewidth}
    \centering
    \includegraphics[width=\linewidth]{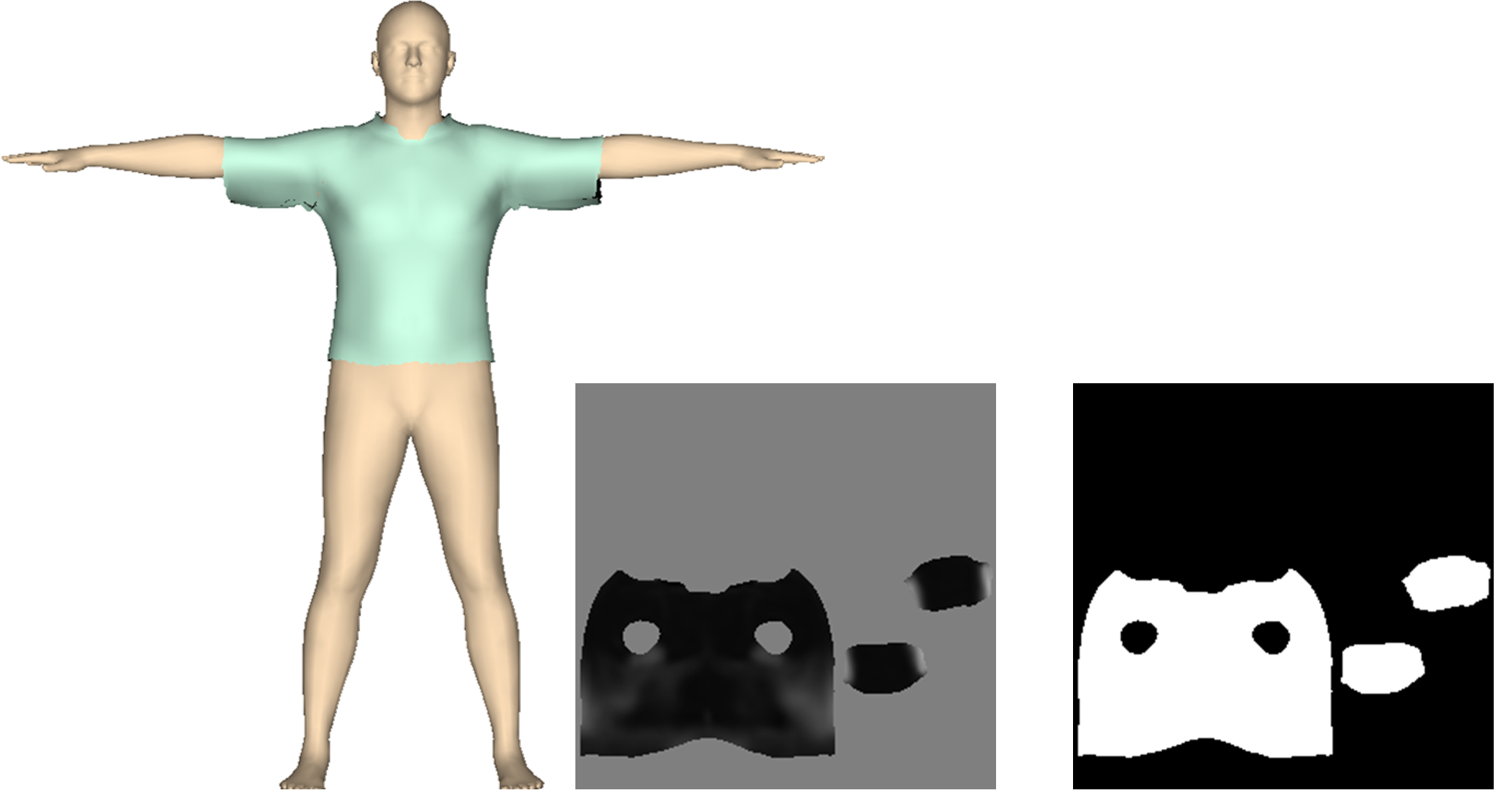}
    \end{minipage}
    }
    \subfigure[]{
    \begin{minipage}[t]{0.47\linewidth}
    \centering
    \includegraphics[width=\linewidth]{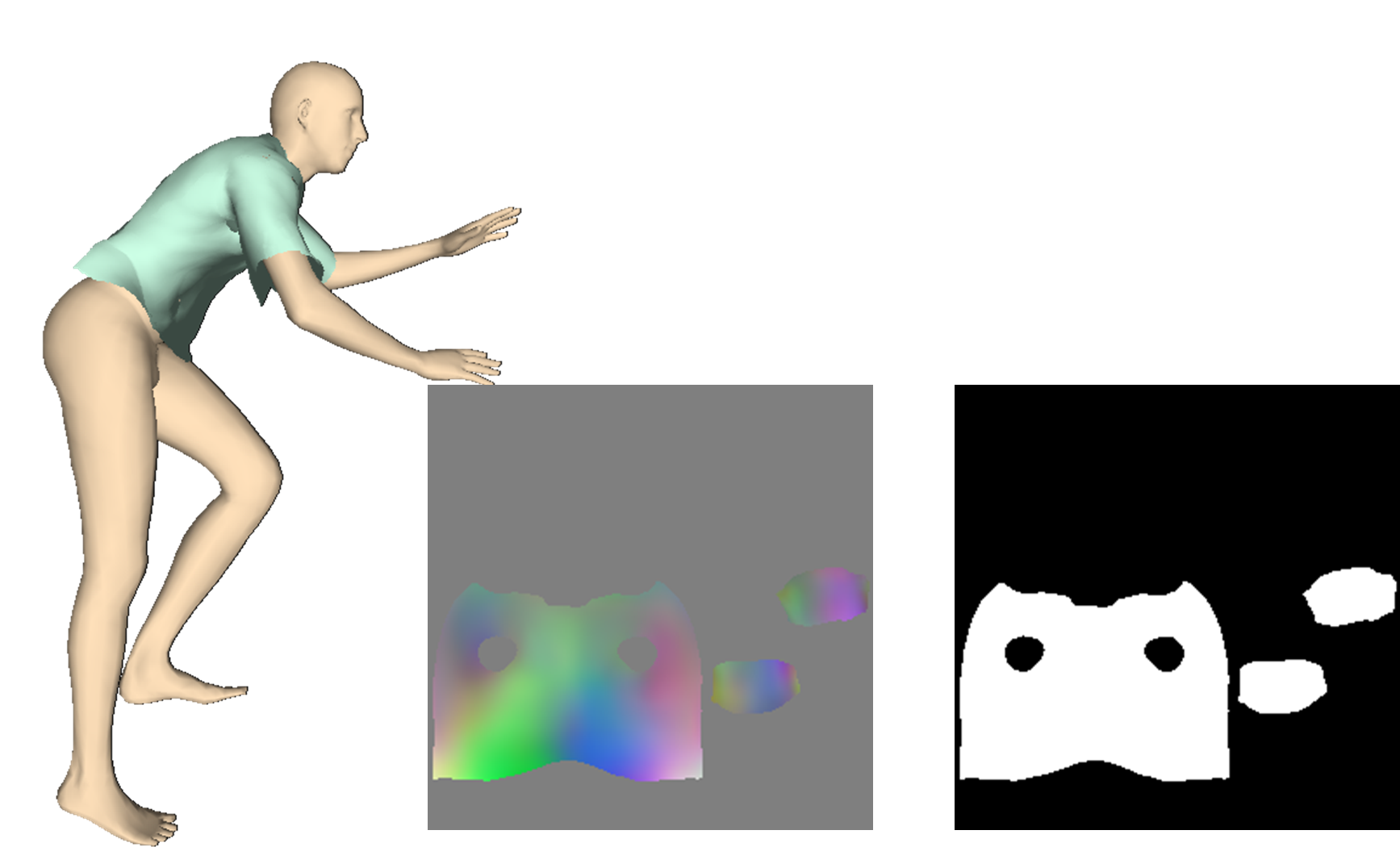}
    \end{minipage}
    }
    \caption{The demonstration on coupling the UV-map representation of the garment. (a) 2D representation for the T-posed garment for Sect.~\ref{subsec:garment_repre}. (b) 2D representation for the randomly posed garment for Sect.~\ref{subsec:garment_anim}.}
    \label{fig:2drepre}
\end{figure}

Specifically, our goal is to find the correspondences between the garment vertices and the human model UV coordinates. Here, we use the same UV map used in ~\cite{VideoBased3DPeople_Alldieck_2018_CVPR}. 
We emit rays from the T-posed SMPL surface that intersect the garment mesh with garment vertices, thus establishing a one-to-one mapping from garment vertices to SMPL surface points.
We denote such an SMPL surface point as the corresponding sub-vertex $\Vec{v}^T_i$ of garment vertex $\Vec{g}^T_i$. In this way, with the predefined UV coordinates of each SMPL triangle face by ~\cite{VideoBased3DPeople_Alldieck_2018_CVPR}, we can accordingly find the UV coordinate $t_i$ of the sub-vertex $\Vec{v}^T_i$, which serves as the corresponding UV coordinate for $\Vec{g}^T_i$. After calculating the length from $\Vec{v}^T_i$ to $\Vec{g}^T_i$, we set the length value as the rendered texel, which indicates the normal distance from $\Vec{v}^T_i$. The rendered T-posed UV-map for a T-shirt is shown in Fig.~\ref{fig:2drepre}(a).


In this way, as shown in Fig.~\ref{fig:2drepre}(a), each garment type is represented into one UV space. For each type of garment (upper garments, pants, dresses), the mask of its UV-map denotes its covering area of the human body and thus contains the topology information of different shape styles. 
The next step is to perform UV map encoding, so that garments with different shape styles and topologies can be then applied to a shape transition framework.

\section{Learning the Garment Shape and Style Space}
\label{subsec:garment_param}

Based on the UV-mask garment representation, we can transfer the complicated 3D shape encoding problem into 2D space by using 2D image auto-encoders.
To achieve garment shape and style transition, the UV-position-based garment representation should be mapped into a continuous space, so that the garment shapes and topologies can be smoothly transitioned. By dimensionality reduction and feature extraction, we map the garment representation UV-map to a low-dimensional feature vector, where both the UV-map and its mask information are encoded into a continuous feature space. Therefore, by editing, interpolating and decoding from the feature space, we can achieve our goal of continuous garment shape transition between different garment topologies and shape styles.

We introduce \ParamNet, a CNN-based network for garment shape and style space learning. The main idea is to leverage a CNN encoder-decoder structure to encode the given T-posed garment UV representation generated in Sect.~\ref{subsec:garment_repre}. Our UV representation contains two pieces of information: (1) the mask, i.e., the area where the UV map has rendered texels illustrates the area where the garment covers the human body (with T-shirts and pants) or the height range of the T-posed dresses; (2) the rendered texels of the UV map illustrate the vertex positions of the garment. Therefore, when performing the encoding, we also make the decoder generate two maps, one for the mask, and the other for the vertex offsets.

\begin{figure}[ht]
    \centering
    \includegraphics[width=0.48\textwidth]{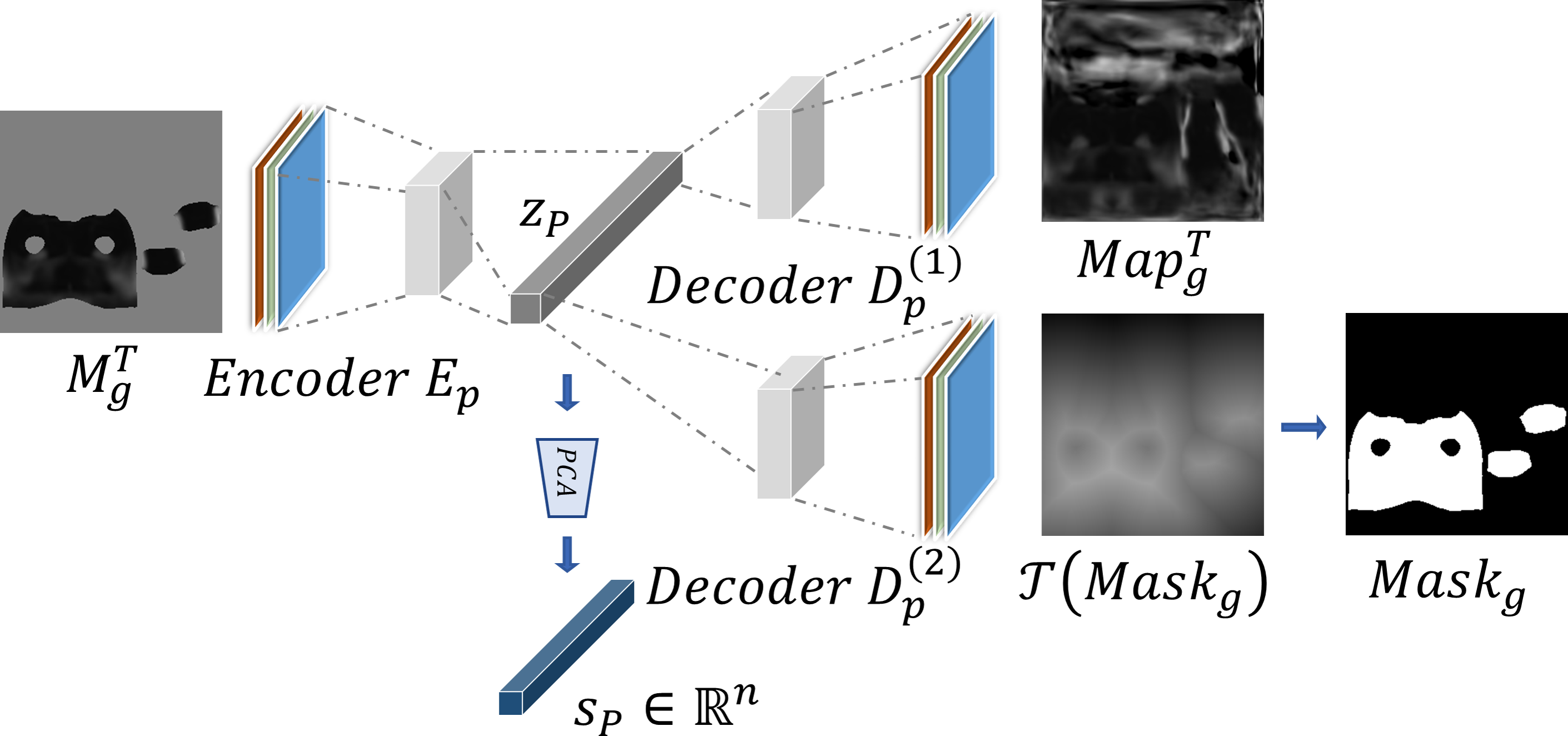}
    \caption{The basic structure of our proposed \ParamNet, which encodes our UV-based garment representation into garment latent space.}
    \label{fig:paramnet}
\end{figure}

As shown in Fig.~\ref{fig:paramnet}, the basic structure of \ParamNet contains two parts, the encoder $E_P(M_g^T)=z_P\in \mathbb{R}^{N}$ encodes the T-posed garment 2D representation $M_g^T$ to a high-dimensional hidden space, and decoder $D_P^{1,2}(z_P)=Map_g^T, Mask_g$ decodes vector $z_P$ in a high-dimensional hidden space to the corresponding map and mask.

\begin{figure}[ht]
    \centering
    \subfigure[]{
    \begin{minipage}[t]{0.95\linewidth}
    \centering
    \includegraphics[width=\linewidth]{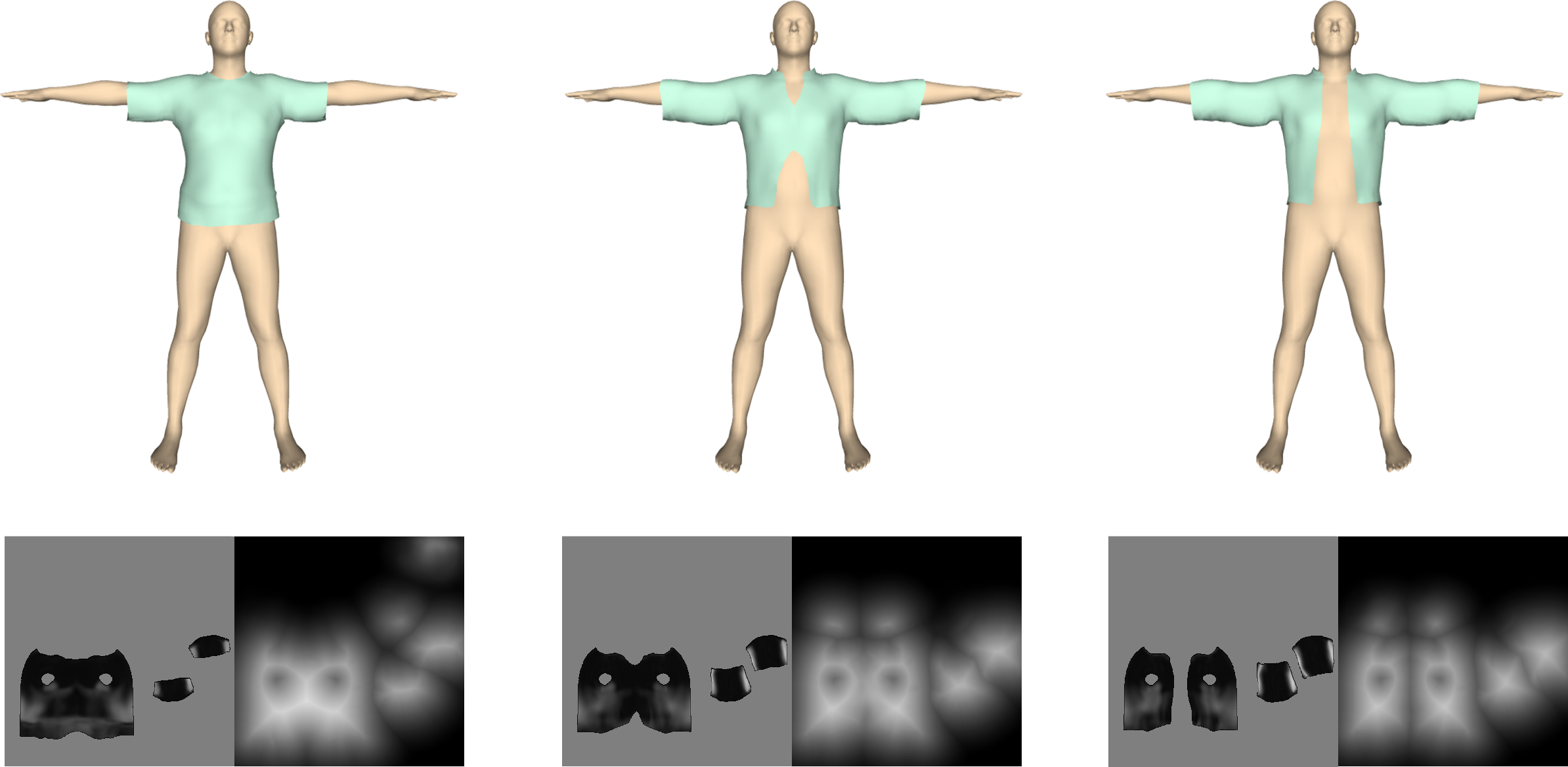}
    \end{minipage}
    }
    \subfigure[]{
    \begin{minipage}[t]{0.95\linewidth}
    \centering
    \includegraphics[width=\linewidth]{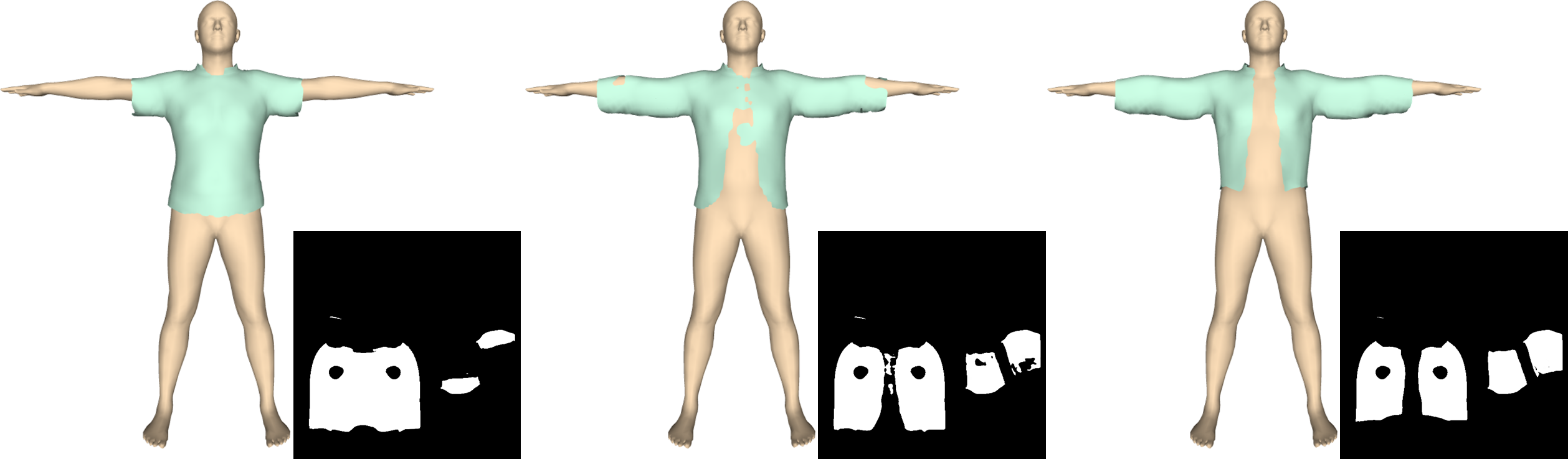}
    \end{minipage}
    }
    \caption{The demonstration of garment shape transition under UV-mask representations: (a) 
    garment shape transition with the proposed distance-transformed masks, (b) garment shape transition with binary masks. Note that without the distance-transform operation, there will be an unnatural transition between different garment shapes.}
    \label{fig:eval_dt}
\end{figure}

In practice, we found that the binary masks could barely perform smooth transitions. This is because the discrete binary mask does not have natural continuity in transition. Therefore, to transform the binary masks into a continuous representation, we propose a distance-transform-based method for pre-processing the masks. Specifically, with a given $Mask_g$, we first generate its \say{bi-distance transform} map as follows:
\begin{equation}
    \mathcal{T}(Mask_g)=\mathcal{DT}(Mask_g)-\mathcal{DT}(\mathcal{I}-Mask_g)
\end{equation}
Here $\mathcal{I}$ is the map with the value 1, and $\mathcal{DT}$ refers to the standard distance transform operation on the mask. The transformed mask, as shown in Fig.~\ref{fig:paramnet}, demonstrates how far a pixel is from the map boundary, and the continuation of the transformed mask makes it easy to be parameterized and learned from the decoder. Therefore, the decoder becomes $D_P^{1,2}(z_P)=Map_g^T, \mathcal{T}(Mask_g)$, and the loss functions are as follows:
\begin{equation}
\begin{aligned}
    \mathcal{L}_P^{(map)} &= ||M_g^T-Map_g^T*Mask_g^{(gt)}||_1 \\
    \mathcal{L}_P^{(mask)} &= ||\mathcal{T}(Mask_g)-\mathcal{T}(Mask_g^{(gt)})||_1
    \label{eq:loss_param}
\end{aligned}
\end{equation}

After the training phase of \ParamNet, the vectors in high-dimensional hidden space $z_P=E_P(M_g^T)$ encrypt the shape variations and characteristics of the T-posed garment shape. To extract the features from the high-dimensional hidden space, we compute the PCA subspace from the hidden space, and sample shape parameters $s_P=E_P(M_g^T)\in \mathbb{R}^{n}$ from the PCA subspace. To recover the T-posed garment shape from shape parameters, we reversely obtain the vector $z_P=PCA^{-1}(s_P)$, perform the decoder operations to obtain the 2D representation $M_g$, and finally generate the garment mesh from it.


The demonstration of our T-posed garment shape transition is shown in Fig.~\ref{fig:eval_dt}(a) and Fig.~\ref{fig:exp_param}, which shows that we can perform a smooth shape transition from short-sleeve T-shirts to long-sleeve shirts, or from skirts to long dresses by interpolating the shape parameters in the feature space. Benefiting from our UV-based representation with mask transformation, we guarantee the continuity in the transition process. The results also demonstrate the function of a general and flexible garment shape encoding and control framework. Although such editing is not strictly semantic (similar to the SMPL~\cite{SMPL:2015} model, which cannot control the shape of a specific body part), different PCA basis can still enable garment shape changes on different dimensions of a garment, as demonstrated in our video demo (1:34-1:54). Fig.~\ref{fig:eval_dt} shows an ablation study using the continuous $\mathcal{DT}$ operations; please refer to Sect.~\ref{sec:results} for more details.


\section{Garment Shape Inference}
\label{subsec:garment_infer}

\label{sec:infer}
\begin{figure}[ht]
    \centering
    \subfigure[]{
    \begin{minipage}[t]{0.20\linewidth}
    \centering
    \includegraphics[width=\linewidth]{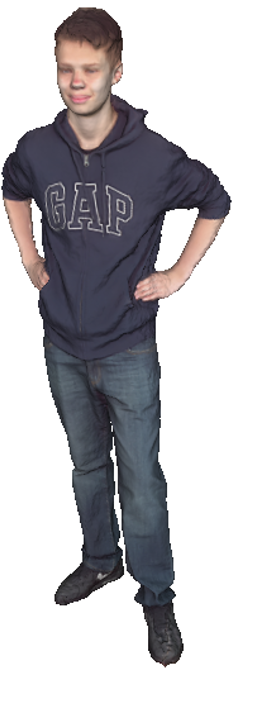}
    \end{minipage}
    }
    \subfigure[]{
    \begin{minipage}[t]{0.20\linewidth}
    \centering
    \includegraphics[width=\linewidth]{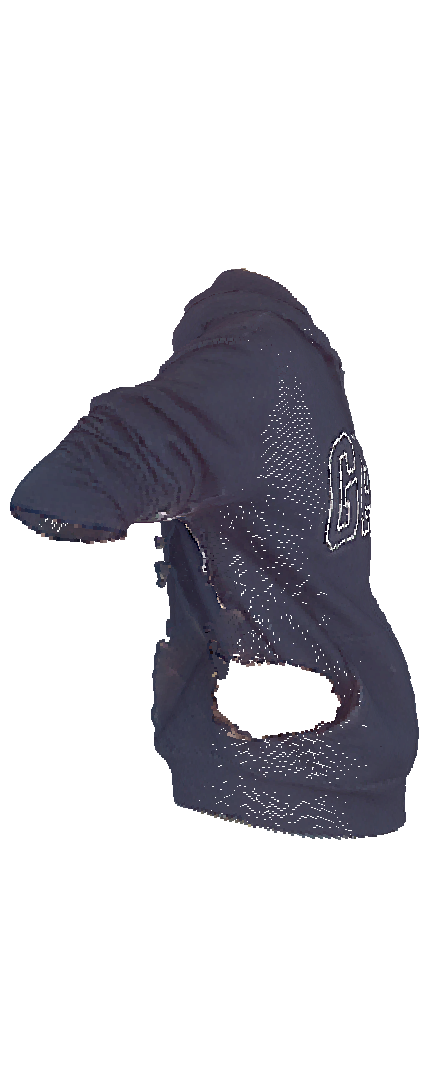}
    \end{minipage}
    }
    \subfigure[]{
    \begin{minipage}[t]{0.24\linewidth}
    \centering
    \includegraphics[width=\linewidth]{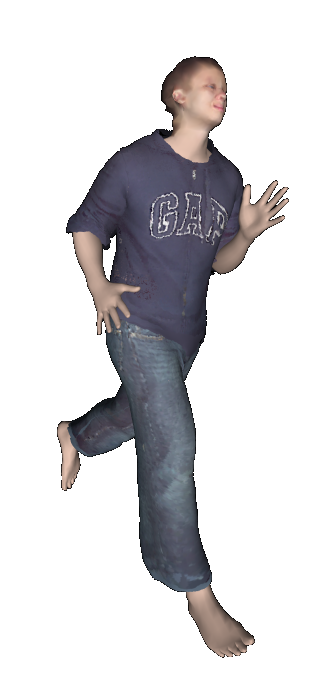}
    \end{minipage}
    }
    \subfigure[]{
    \begin{minipage}[t]{0.24\linewidth}
    \centering
    \includegraphics[width=\linewidth]{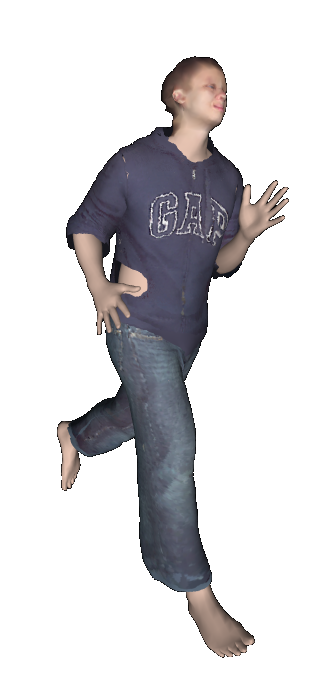}
    \end{minipage}
    }
    \caption{The demonstration of different garment shape inference methods. From left to right: (a) input 3D scan, (b) segmented garment, (c) garment animation result with garment shape inferred by \InferNet, (d) garment animation result with garment shape directly obtained from the scan. }
    \label{fig:eval_InferNet}
\end{figure}

Based on our proposed garment modeling framework and continuous UV-mask representation, a PointNet-based \InferNet is proposed to map the 3D garment mesh data domain onto the proposed garment UV space.
Note that an alternative method for garment shape inference given a 3D garment scan is directly obtaining the corresponding UV/mask-maps from the scan, similar to our dataset preparation method. However, this method is not feasible. First, for cases when the garment scan is not complete (e.g., Fig.~\ref{fig:eval_InferNet}(a)(b), the waist of the garment was partially occluded by the hands of the subject), the corresponding UV-maps cannot be generated properly. Second, the garment scans need to be deformed to the standard T-pose for animation, which will produce skinning artifacts especially on the human underarm area.
To solve these problems, we propose our garment shape inference module, i.e., \InferNet, which maps the scanned point clouds to our garment feature space, and generates garment shapes accordingly.
As shown in Fig.~\ref{fig:eval_InferNet}, our proposed \InferNet is necessary for generating complete garment shapes.

\begin{figure}[ht]
    \centering
    \includegraphics[width=0.48\textwidth]{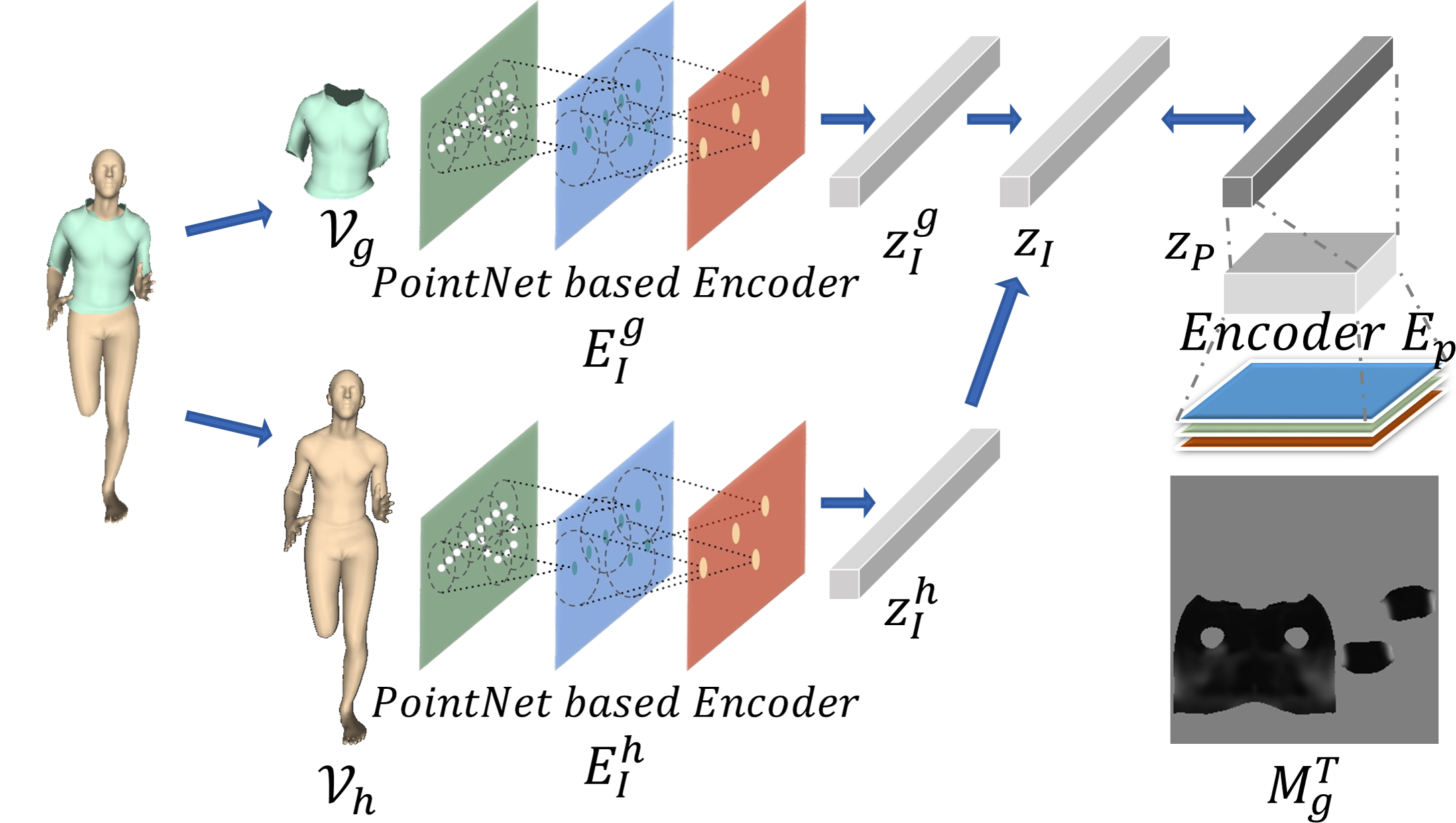}
    \caption{The basic structure of our \InferNet, which infers the garment shape parameters from the input 3D scans..}
    \label{fig:infernet}
\end{figure}


To reconstruct the garment shape from any given 3D raw data, and further perform static-to-dynamic garment 3D animation and 3D editing, we introduce the garment shape inference module, which takes a given garment mesh under randomly posed humans as input and extracts the corresponding garment shape parameters. Our method encodes garment shapes with different styles and topologies into a feature space, enabling garment shape extraction from the encoded space.
Benefiting from our UV-mask-based garment representation module, our shape inference module can support different garment shapes and styles. In regard to previous works, SIZER~\cite{SIZER_tiwari20sizer} can only perform static garment editing. TailorNet~\cite{TailorNet_Patel_2020_CVPR} has the potential for shape extraction by un-posing the scan to a standard pose and fitting to its fixed garment template, but it cannot perform large garment shape and topology changes, nor can it deal with garments that do not fit to its templates.

We introduce \InferNet to achieve this task. As shown in Fig.~\ref{fig:infernet}, in our garment inference module, the input of our model is a pre-segmented garment mesh with an aligned human model. 
For a given garment mesh $\mathcal{V}_g$ and the corresponding human model mesh $\mathcal{V}_h$, our goal is to generate the corresponding garment parameters $s_g$. The basic structure of \InferNet contains two-branch PointNet-based encoders $E_I^h(\mathcal{V}_h)=z_I^h$ and $E_I^g(\mathcal{V}_g)=z_I^g$, which separately encode the input randomly posed human mesh and garment mesh to hidden space vectors. We implement a fully connected operator $\mathcal{F}$ for extracting features from $z_I^h$ and $z_I^g$ as $\mathcal{F}(z_I^h,z_I^g)=z_I$. The loss is then introduced to constrain the output feature $z_I$ to have less deviation with the vector $z_g$ encoding the shape parameters (see Sect.~\ref{subsec:garment_param}):
\begin{equation}
    \mathcal{L}_I = ||z_I-z_P||_1
    \label{eq:loss_infer}
\end{equation}


As the PointNet~\cite{PointNet_Qi_2017_CVPR}-based encoder structure does not rely on the topology or vertex numbers of the input garments, with our style-flexible garment shape representation method, we can extract the garment shape parameters from garment meshes with any 3D inputs. The results are shown in Fig.~\ref{fig:teaser}(b) and Fig.~\ref{fig:exp_infer}.


\section{Garment Animation}
\label{subsec:garment_anim}

\begin{figure}[ht]
    \centering
    \includegraphics[width=0.48\textwidth]{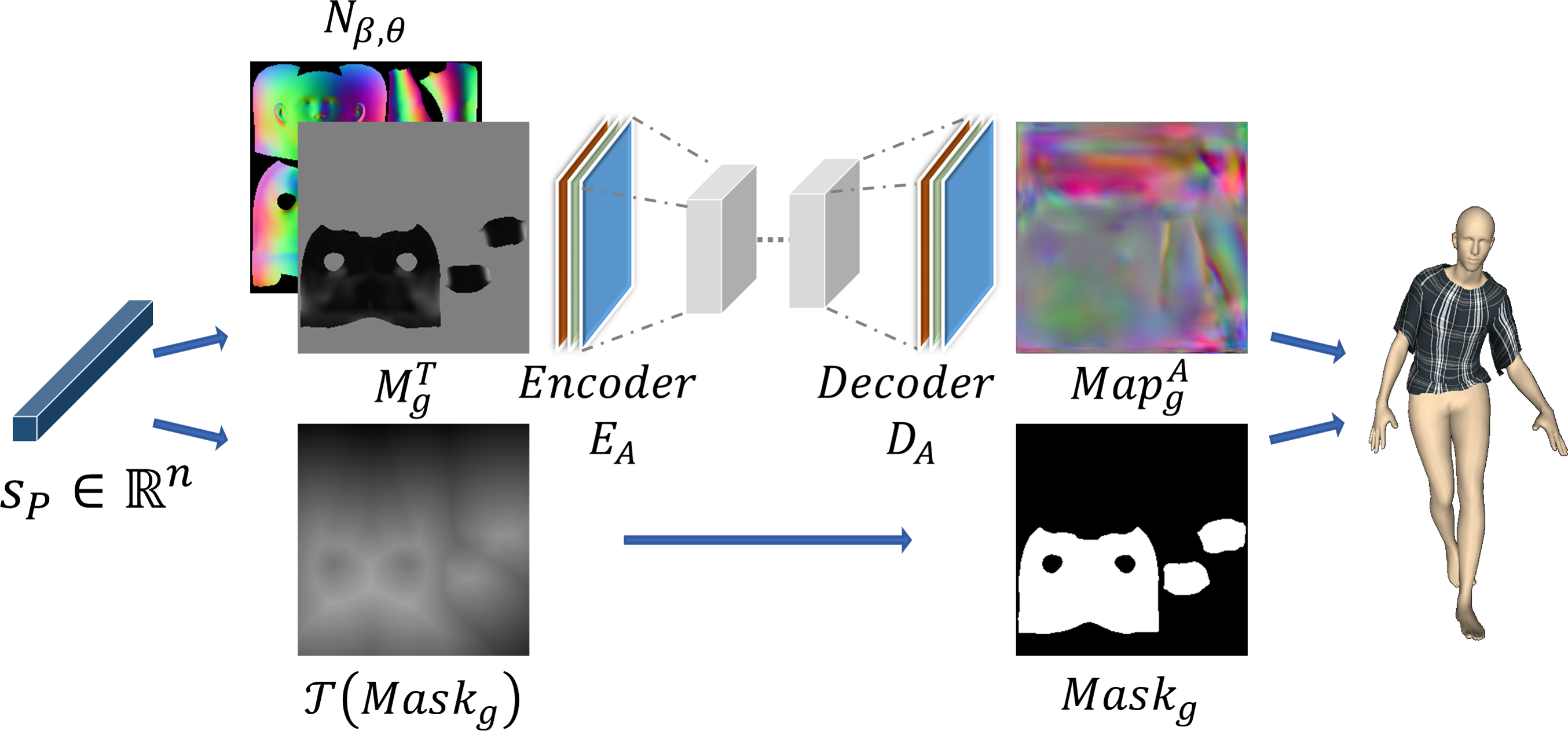}
    \caption{The basic structure of our \AnimNet, which generates garment dynamics under arbitrary human poses and shapes.}
    \label{fig:animnet}
\end{figure}

As proposed in Sect.~\ref{subsec:garment_repre} and Sect.~\ref{subsec:garment_param}, we represent the garment shape with the UV-map and encode it into a feature space. In addition to being applied to the garment shape style transition, the representation and encoding module can also be used for dynamic garment animation to animate the clothed human into arbitrary new poses, which can also be applied to the reconstructed garment shape from Sect.~\ref{subsec:garment_infer}. 
To better formulate the connection between T-posed garments and corresponding garments under arbitrary poses, we fix the correspondence map defined by the T-posed 1-channel normal distance map, and calculate a 3-channel shift map for each posed garment. Therefore, the static and dynamic garment representations are semantically consistent and suitable for further applications such as garment animation. 

Benefiting from our unified UV-mask garment representation, our animation module generates various garment dynamics with different shape styles in a single network, which has not been demonstrated even by the concurrent unified garment representation framework~\cite{SIMPLicit_Corona_2021_CVPR}.
The garment animation module in our \ourpaper takes the input garment shape parameters from Sect.~\ref{subsec:garment_param} with human pose and shape and generates the animated garment mesh. To achieve this goal, we introduce \AnimNet, which is a CNN-based network for the garment animation module. 

The first step is to represent garments under arbitrary poses. As shown in Fig.~\ref{fig:2drepre}, we establish a topology-consistent coupling UV-map for a T-posed garment and the same garment under arbitrary poses. As the previous steps determine the UV coordinates of each garment vertex, for a garment animated on a human with other poses, we fix the UV coordinates and set the rendered texels representing the animated shape. Specifically, 
we calculate the position shift between garment vertex $\Vec{g}^T_i$ and corresponding SMPL sub-vertex $\Vec{v}^T_i$, and set position shift $(dx, dy, dz)$ as the rendered texels. 

There are three main advantages for our coupling UV-map representation. 
First, by fixing the correspondence between garment vertices and SMPL UV coordinates, a one-to-one mapping can be applied to the T-posed garment vertices and animated garment vertices, and randomly posed garments, e.g., floating front-opening T-shirts or folding skirts, can be represented more easily.
Second, with the same UV coordinate for every garment vertex, the mapping between T-posed garments and its randomly posed condition can be learned more easily using a CNN-based network. Third, since the coupling UV-map has the same mask, during animation, we only need to infer the rendered texels of the second map.

We denote $M_g^T$ for the T-posed standard garment UV map, and $M_g^A$ for the animated garment UV map. In \AnimNet, we take $M_g^T$ as input and generate $M_g^A$ as output. Meanwhile, to encode the human pose and shape information, we find that the normal information actually guides the position map of the garment; therefore, we use the normal map $N_{\beta, \theta}$ of the human model to represent the human shape $\beta$ and pose $\theta$ information. As shown in Fig.~\ref{fig:animnet}, we use a CNN-based encoder $E_A$ and decoder $D_A$ with skip connections to generate the inferred garment map $Map_g^A$. The main loss function is as follows:
\begin{equation}
\begin{aligned}
    M_g^A = Map_g^A*Mask_g^{(gt)} \\ 
    \mathcal{L}_A^{(map)} = ||M_g^{A(gt)}-M_g^A||_1
    \label{eq:loss_anim}
\end{aligned}
\end{equation}
Here $Mask_g^{(gt)}$ is the ground truth mask, and we only need to constraint the generated position map to have the same value as the ground truth map inside the masked area, as the input garment shape parameters contain the mask information.

\begin{figure}[ht]
    \centering
    \includegraphics[width=0.40\textwidth]{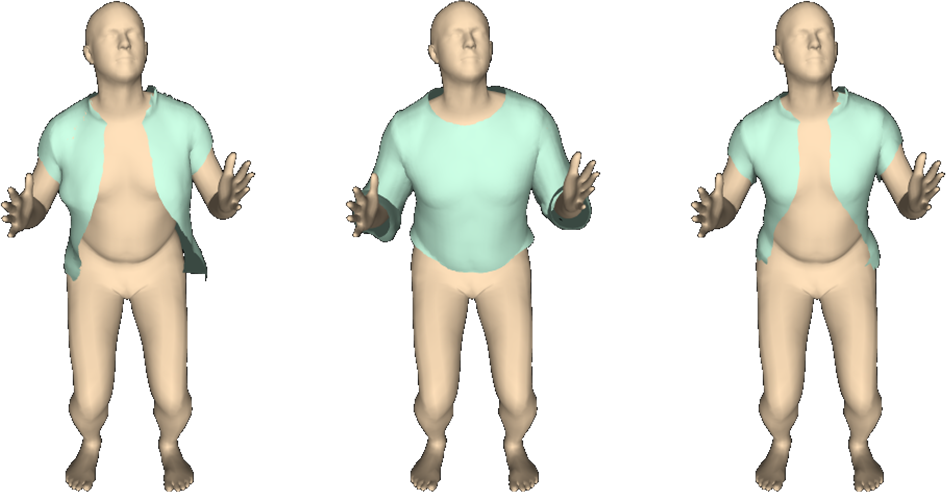}
    \caption{The demonstration of different styles garment animation. From left to right: front-opening T-shirt animation, front-closing T-shirt animation, and skinning results with front-opening T-shirt. The results clarify that \AnimNet learns garment dynamics from the encoded garment styles.}
    \label{fig:eval_mask_garment}
\end{figure}

Note that our UV-mask-based garment representation implicitly encodes the garment shape dynamic information, which helps \AnimNet to learn the deformation of different garment styles when performing garment animation(e.g. front-opening/closing garments). As shown in Fig.~\ref{fig:eval_mask_garment}, the network learns garment dynamic shapes with different garment styles, such as the floating bottom part of the front-opening T-shirt. The UV-masks not only mask the garment geometry, but also encode garment styles, which is reflected in animation results. The example garment animation results are shown in Fig.~\ref{fig:exp_param} and Fig.~\ref{fig:exp_shape}, which show that we can animate different types of garments with various human shapes, poses and garment styles. 


\section{Garment Shape-Style Editing}
\label{subsec:garment_edit}

As mentioned in Sect~\ref{subsec:garment_param}, we determine the PCA subspace from the garment representation latent space, which serves as a more compact and semantic encoding of garment style variation than the original latent space. Therefore, we can perform garment shape and style transition and editing by shifting the PCA space vectors.

Take the garment reconstruction procedure as an example. Given garment 3D scans and human models, with the garment shape parameters solved by \InferNet, we can perform shape editing by the following steps: (1) mapping the shape parameters to the same PCA subspace calculated in Sect.~\ref{subsec:garment_param} as $s_I=PCA^g(z_I)$, (2) editing some dimensions of shape parameters $s_I$ as $s_I'$, (3) obtaining the vector $z_I'=PCA^{-1}(s_I')$, and (4) recovering a new garment shape from $z_I'$. 

\begin{figure*}[ht]
    \centering
    \subfigure[]{
    \begin{minipage}[t]{0.60\linewidth}
    \centering
    \includegraphics[width=\linewidth]{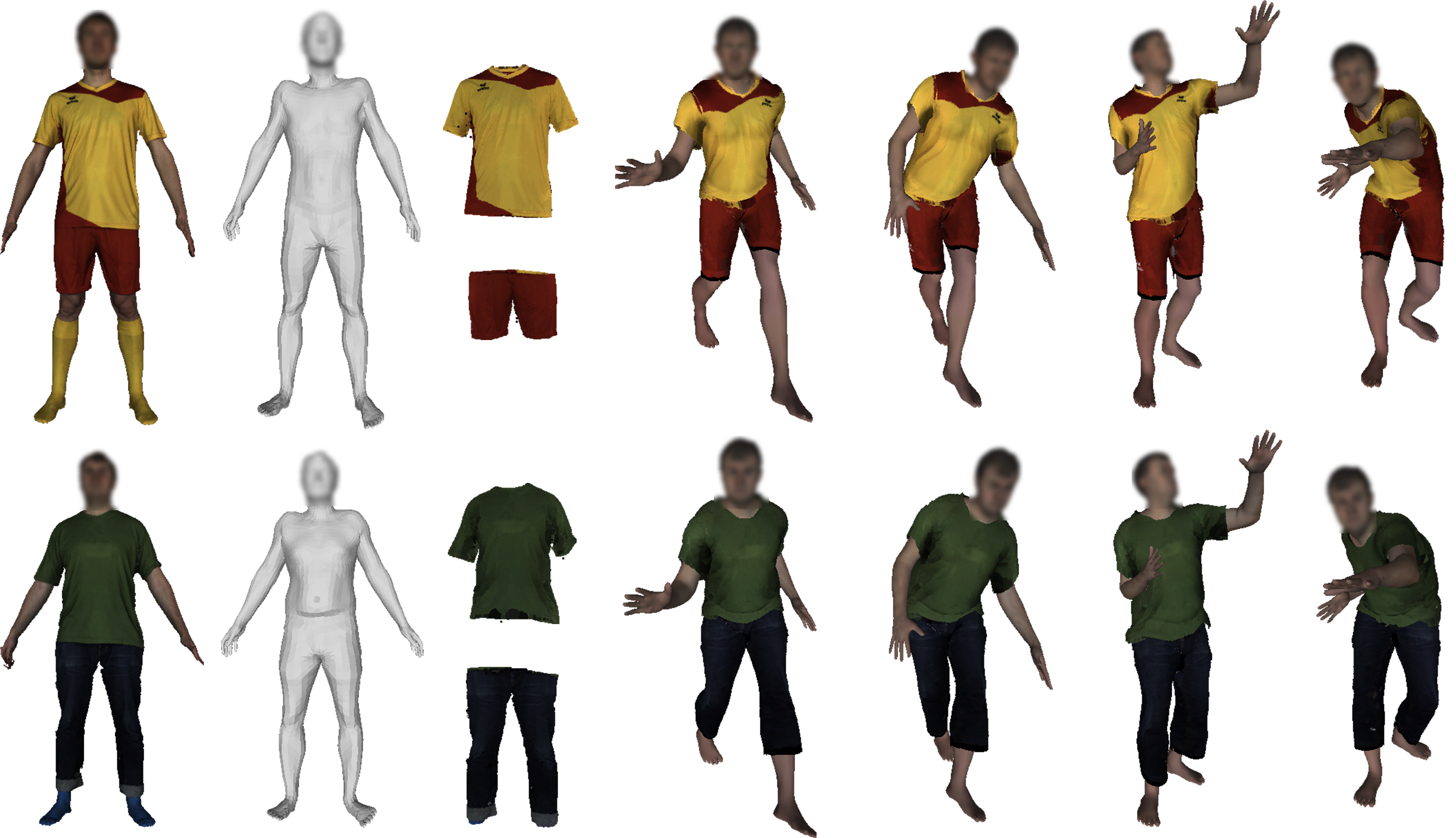}
    \end{minipage}
    }
    \subfigure[]{
    \begin{minipage}[t]{0.365\linewidth}
    \centering
    \includegraphics[width=\linewidth]{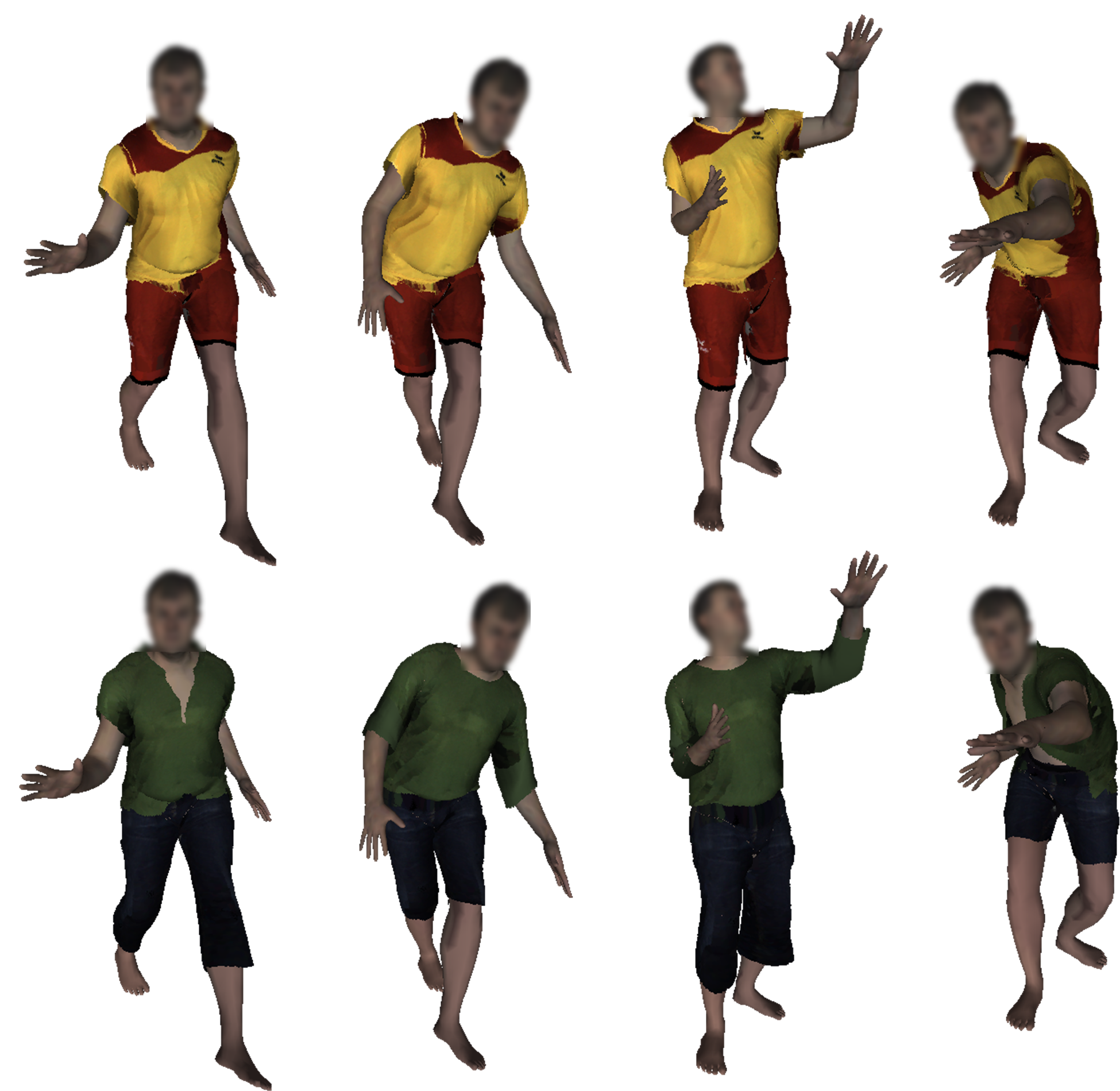}
    \end{minipage}
    }
    \caption{The experimental results of our garment inference module. (a) Garment inference with a given input, from left to right: original 3D scan, aligned human model, 3D segmented T-shirts with pants, and animation results. (b) Top: garment retargeting results, bottom: garment 3D editing results.}
    \label{fig:exp_infer}
\end{figure*}

After the garment editing step, the new garment can then be fed to our \AnimNet for garment animation, so that dynamic garment shape editing can be applied to the given pose sequences. The results are shown in Fig.~\ref{fig:teaser}(c)(d)(e) and Fig.~\ref{fig:exp_infer}(b). 

\begin{figure}[ht]
    \centering
    \includegraphics[width=0.45\textwidth]{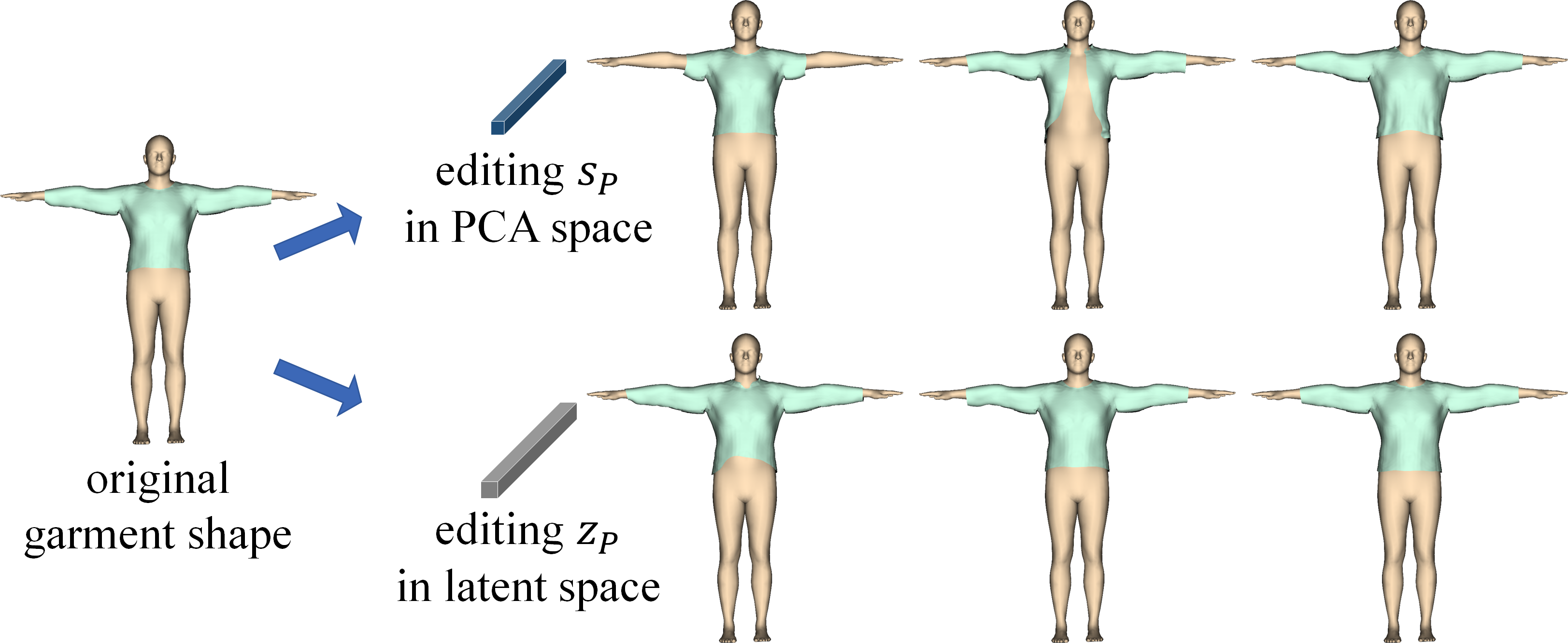}
    \caption{The ablation study of garment editing in PCA space and the original latent space. Top: editing parameters in PCA space; bottom: editing parameters in original latent space. Different garment shape styles can be edited easily by shifting the PCA vectors. In contrast, directly editing the original latent space vectors can hardly yield meaningful garment shape editing results. }
    \label{fig:major1_PCA}
\end{figure}

To show the necessity of PCA for garment editing, we perform an ablation study for editing the garment parameters in two ways: (1) editing one dimension each time in PCA space, and (2) editing one dimension each time in the original latent space vector, with the relative parameter shift degree. As shown in Fig.~\ref{fig:major1_PCA}, we can perform convenient and semantic garment editing by shifting different dimensions in PCA space, while such editing can hardly be performed by editing in the original latent space. This is because the PCA operation encodes the original garment shape space in a more compact form and extracts the semantic style patterns. 


\section{Garments not homotopy to human body}

\begin{figure}[ht]
    \centering
    \subfigure[]{
    \begin{minipage}[t]{0.47\linewidth}
    \centering
    \includegraphics[width=\linewidth]{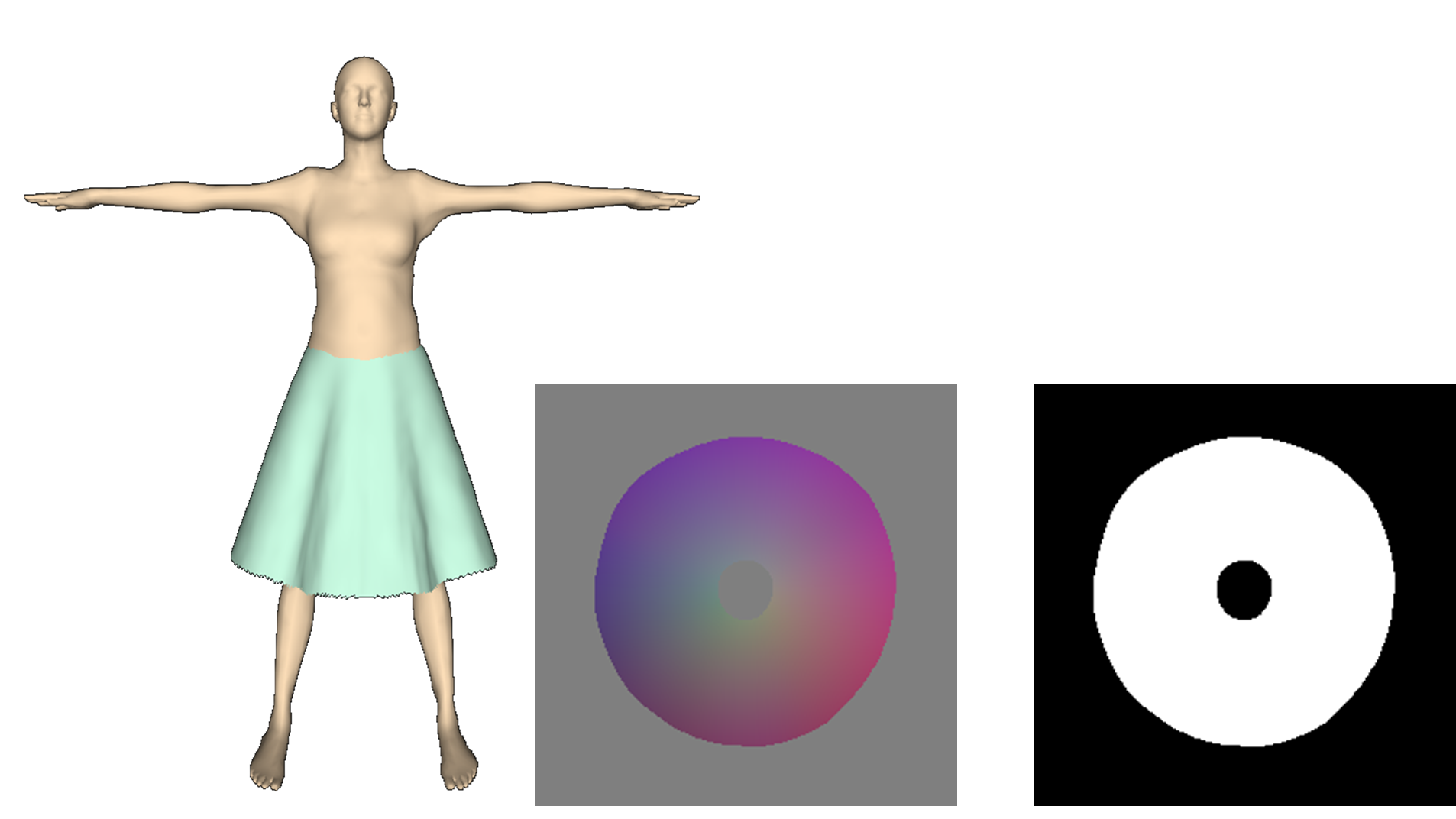}
    \end{minipage}
    }
    \subfigure[]{
    \begin{minipage}[t]{0.47\linewidth}
    \centering
    \includegraphics[width=\linewidth]{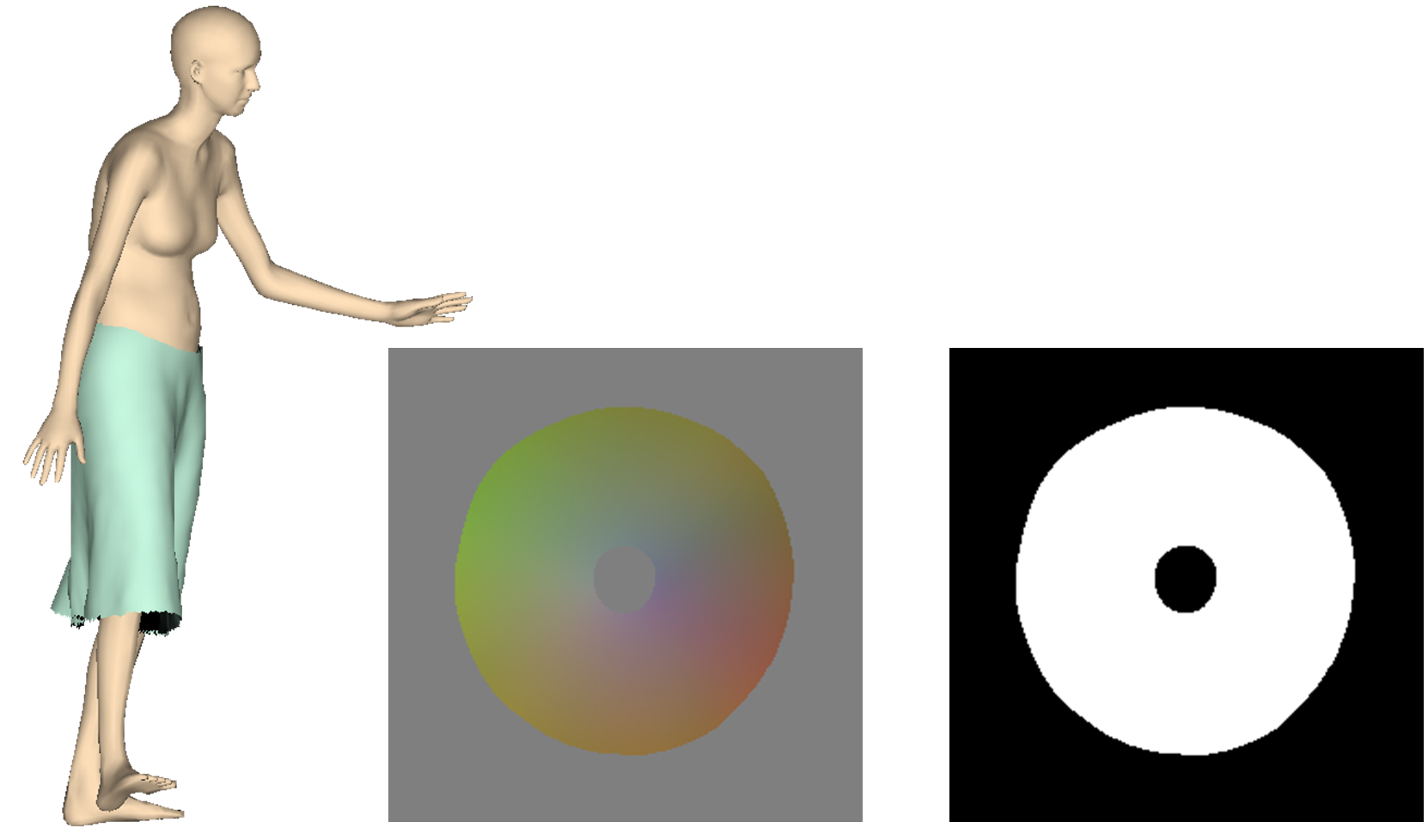}
    \end{minipage}
    }
    \caption{The demonstration on coupling the UV-map representation of the garment that is not homotopy to the human body. (a) 2D representation for the T-posed skirt. (b) 2D representation for the randomly posed skirt for Sect.~\ref{subsec:garment_anim}.}
    \label{fig:2drepre_skirt}
\end{figure}

Garments that are not homotopy to the human body, e.g., skirts and dresses, are not suitable for our SMPL-UV-based representation. Therefore, we separately design the UV-mask representation for those garments. We map them to an independent UV coordinate to better reflect the characteristics of the garment geometry. The boundary of the UV map demonstrates the edges and basic shape information (such as the height of dresses), and the rendered texels indicate the 3D positions of the garment vertices.

\begin{figure}[ht]
    \centering
    \includegraphics[width=0.45\textwidth]{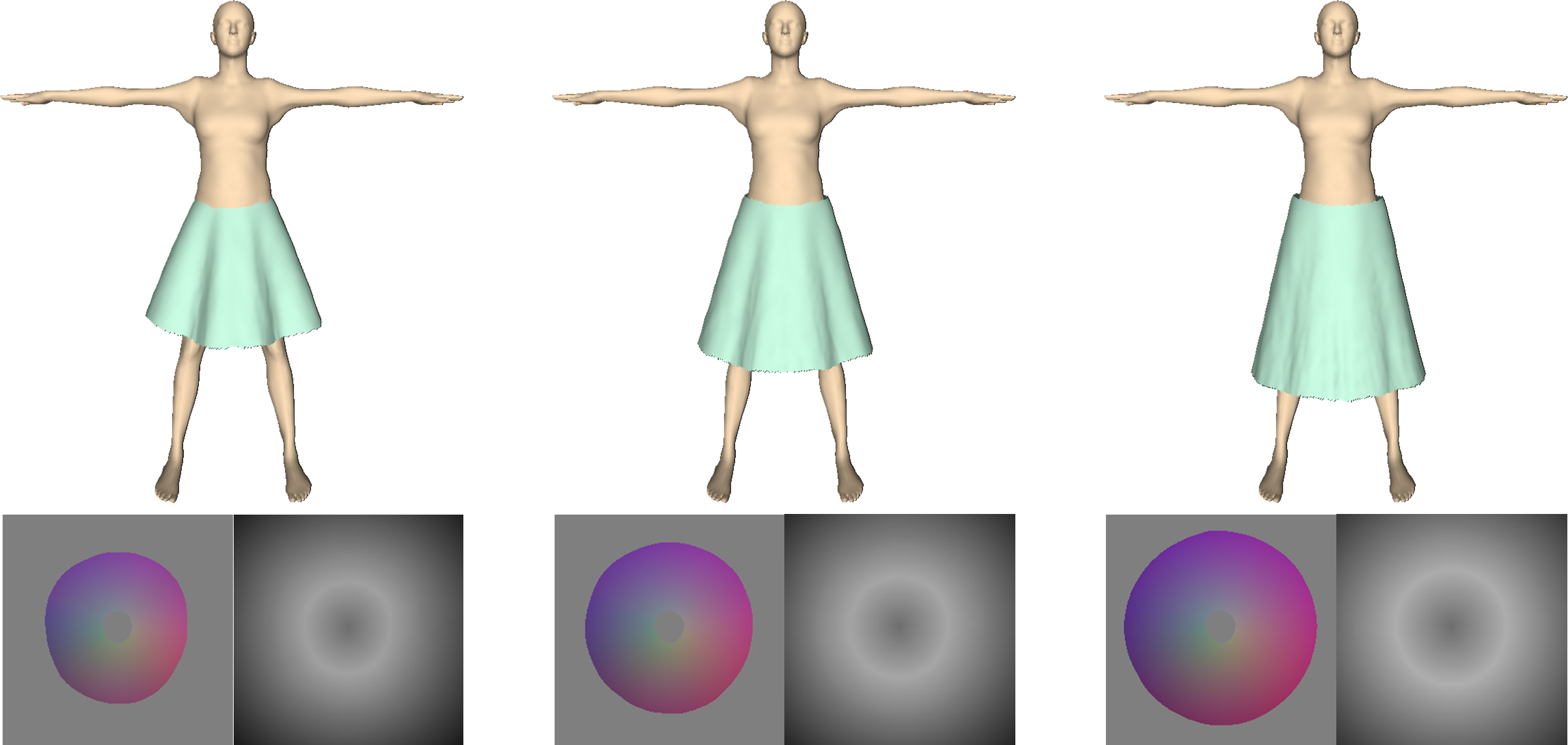}
    \caption{The demonstration of the skirt shape transition under UV-mask representations}
    \label{fig:eval_dt_skirt}
\end{figure}

Specifically, when dealing with clothing that is not homotopy to the human surface (e.g., dresses), we set an independent UV coordinate accordingly. For T-posed dresses and skirts, as their geometry circles around the lower body, we leverage cylindrical coordinates and calculate the UV coordinates as follows: for each garment vertex $\Vec{g}'^T_i$, we transfer it into cylindrical coordinates: $\Vec{g}'^T_i(x,y,z) \rightarrow \Vec{g}'^T_i(r,y,\theta)$ where $r=\sqrt{x^2+z^2}$ and $\theta=\arctan(z,x)$. The UV coordinate $t'_i$ for $\Vec{g}'^T_i(r,y,\theta)$ is $t'_i=((y_0-y)\cos(\theta)+0.5, (y_0-y)\sin(\theta)+0.5)$, and the rendered texel is just $(x, y, z)$ to indicate the vertex positions. Here, $y_0$ serves as the height threshold of the skirts; in practice, we set $y_0=0.2$ above the root joint of the human model. The rendered T-posed UV-map for a skirt is shown in Fig.~\ref{fig:2drepre_skirt}(a). Additionally, continuous style transitions can be performed similar to Sect.~\ref{subsec:garment_param}, and the results are shown in Fig.~\ref{fig:eval_dt_skirt}.

For garments under arbitrary poses, similar to our procedure in Sect.~\ref{subsec:garment_anim}, we fix the garment vertex UV coordinates, and directly use the vertex positions to set the rendered texels, as used in the T-posed scenario. The rendered arbitrarily posed UV-map for a skirt is shown in Fig.~\ref{fig:2drepre_skirt}(b). 

\begin{figure*}[t]
    \centering
    \includegraphics[width=0.95\linewidth]{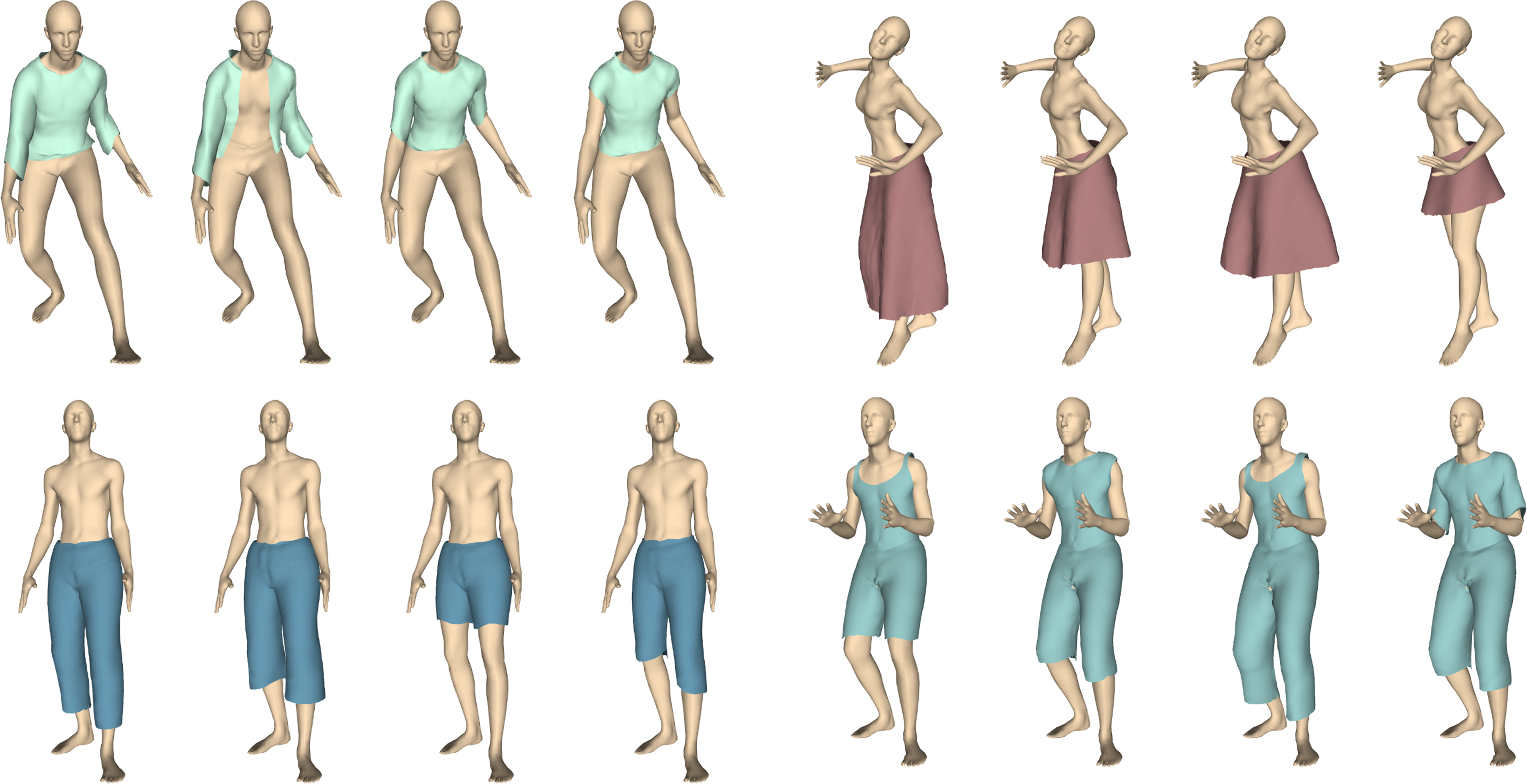}
    \caption{The demonstration of our shape variation for different kinds of garments. Each block shows the garment shape parameter variation on one type of garments.}
    \label{fig:exp_param}
\end{figure*}

\begin{figure*}[ht]
    \centering
    \includegraphics[width=0.95\linewidth]{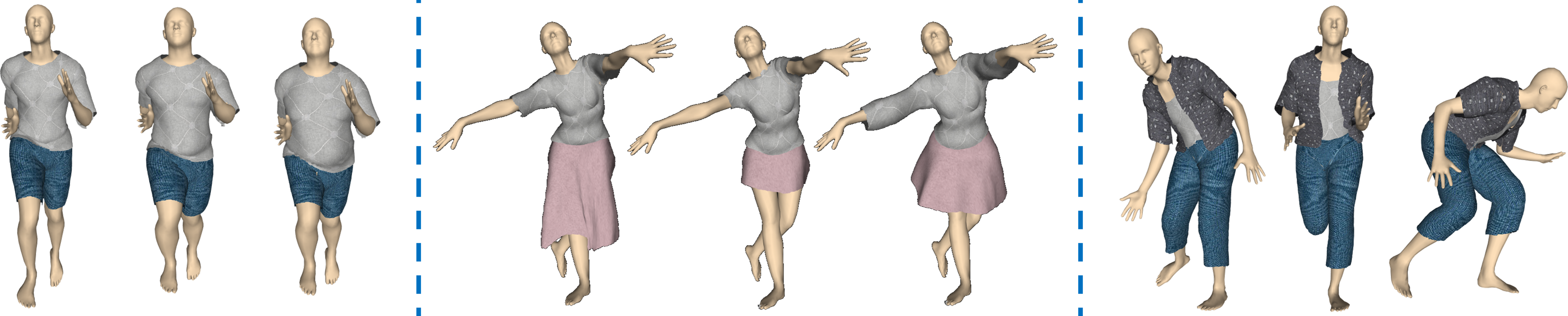}
    \caption{The demonstration of garment animation results. Left: the same garments under different human shapes. Middle: different garment styles on the same person. Right: the same garments under different human poses.}
    \label{fig:exp_shape}
\end{figure*}

Apart from the different UV-coordinate layouts of different types of garments, the main pipeline (\ParamNet, \AnimNet and \InferNet) works the same as garments homotopy to human bodies. As shown in Fig.~\ref{fig:exp_param}, Fig.~\ref{fig:exp_shape} and our video demo, we can perform skirt style editing with robust and vivid animation results, which shows the robustness of our UV-mask-based representation pipeline for dealing with different types of garments.

\section{Experiments}
\label{sec:results}

In our experiments, we use CLOTH3D~\cite{CLOTH3D_2020_ECCV}, which is a large-scale synthetic dataset with various garment shape styles suitable for our pipeline for our training and testing data. We test our garment encoding module with four garment kinds: upper garments, pants, skirts, and jumpsuits. The animation module is tested with all these kinds of garments. In addition, we take the BUFF Dataset~\cite{BUFF_Zhang_2017_CVPR} and Twindom Dataset (https://web.twindom.com/) as input to test our garment inference module for 3D garment animation and editing from input 3D data.

The network structures for our CNN-based encoders, e.g., $E_P$ in \ParamNet (Sect.~5) and $E_A$ in \AnimNet (Sect.~6), are based on the ResNet-18~\cite{ResNet_He_2016_CVPR} structure. The decoders accordingly are six stacked up-sampling layers with convolution layers. The structure for our PointNet-based~\cite{PointNet_Qi_2017_CVPR} encoders $E_I^g$ and $E_I^h$ (Sect.~7) is based on PointNet structures for extracting features from point clouds.

In addition to the data preparation, with the NVIDIA GeForce GTX TITAN X GPU, the training procedure for \ParamNet takes approximately 50 hours, while \AnimNet and \InferNet take 100 hours separately for each kind of garment. The garment rendering results generated from the network output, together with a standard collision resolving procedure, take approximately 2 seconds per human per frame with one garment.


\textbf{Garment shape representation and animation.} To evaluate our garment shape representation results, Fig.~\ref{fig:exp_param} demonstrates the garment shape variations controlled by different parameters, with the PCA parameter $s_P$ varying within the range of 1.0 $\sigma$. The results in Fig.~\ref{fig:exp_param} show that we can perform plausible garment shape variation from long-sleeve shirts to short-sleeve T-shirts, from front-closing T-shirts to front-opening T-shirts, from long pants to shorts, and from long dresses to short skirts, which clarifies that our method provides a more general garment shape representation model than TailorNet~\cite{TailorNet_Patel_2020_CVPR}. Additionally, as demonstrated in Fig.~\ref{fig:exp_shape}, given different garment styles and different body shapes, we can generate clothed human animation results, which makes our framework capable of representing garment shapes under various human shapes, poses and garment styles.

To demonstrate the necessity of the $\mathcal{DT}$ operation, we evaluate garment shape transition with and without the \say{bi-distance transform} operation. To directly use binary masks, we use a network structure for garment shape representation similar to \ParamNet. As shown in Fig.~\ref{fig:eval_dt}, the lack of a continuous boundary constraint leads to an unnatural shape transition. In contrast, with our \say{bi-distance transform}, the masks are transformed smoothly, thus generating continuous garment shape transition results.

\begin{figure}[ht]
    \centering
    \includegraphics[width=0.95\linewidth]{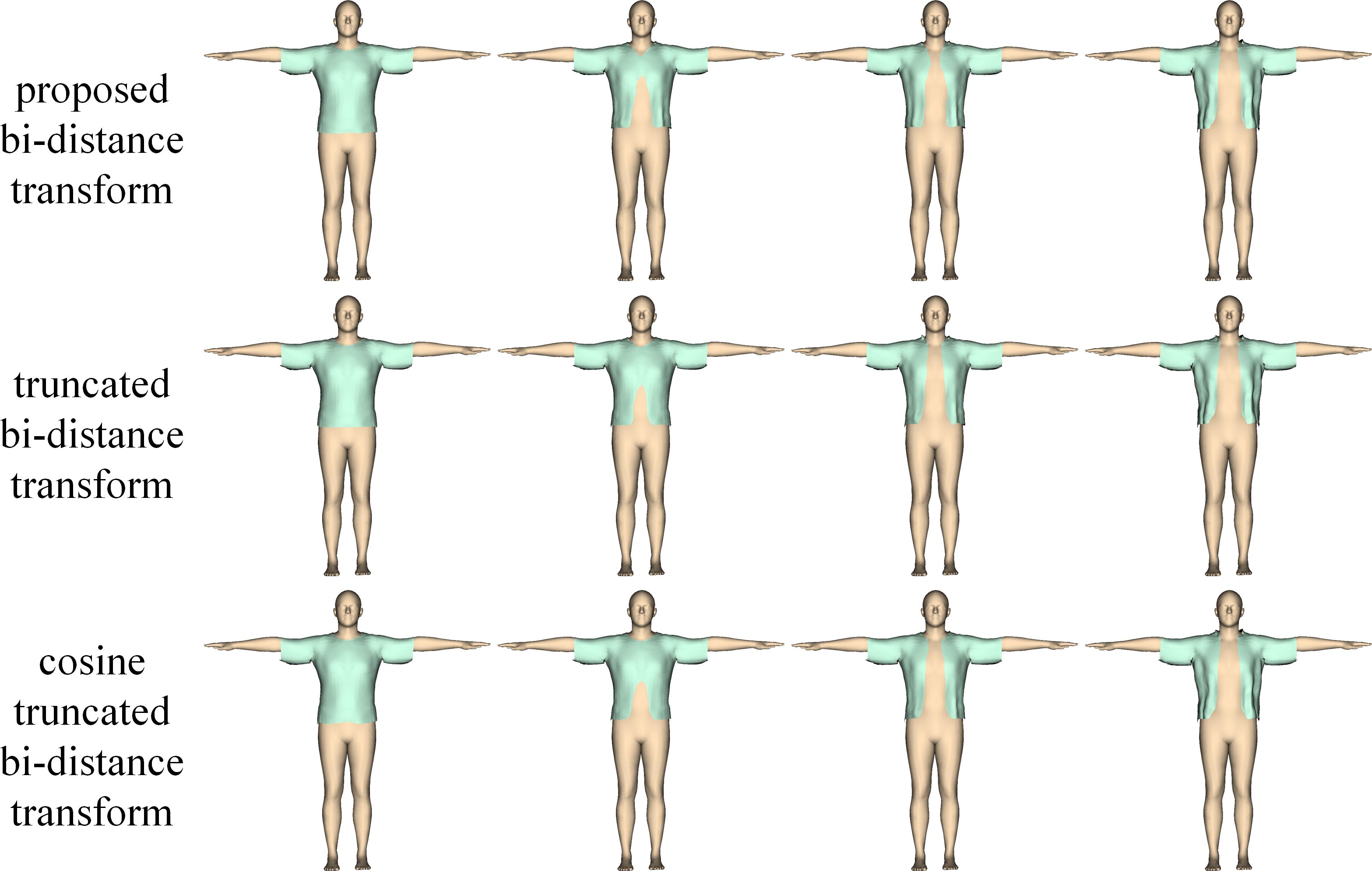}
    \caption{The ablation study on smooth garment transition using different continuous mask transforms based on the proposed bi-distance transform. From top to bottom: garment shape transition using the proposed method, the truncated bi-distance transform and the cosine encoding on the bi-distance transform.}
    \label{fig:major1_DT}
\end{figure}

It should be clarified that although the $\mathcal{DT}$ operations change the binary masks into a continuous form, other continuous $\mathcal{DT}$ transforms, such as the truncated distance transform or cosine encoding based on the distance transform images, can also be used. As shown in Fig.~\ref{fig:major1_DT}, we show that the continuous transforms based on our bi-distance transform can also be applied for a smooth transition.

\begin{figure}[ht]
    \centering
    \includegraphics[width=0.9\linewidth]{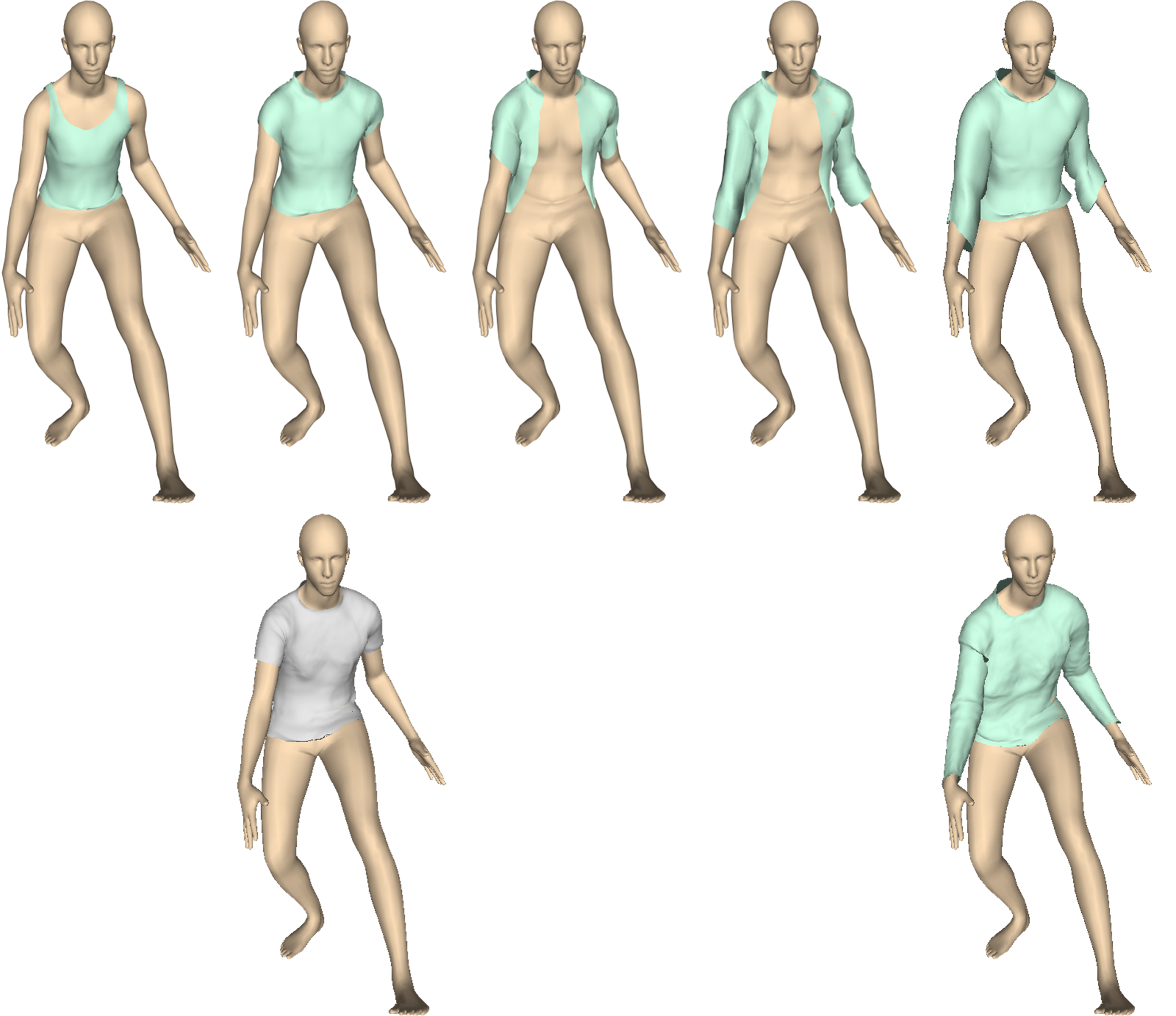}
    \caption{Comparison between our method and TailorNet~\cite{TailorNet_Patel_2020_CVPR} on garment representation. Top: our animated garment shapes with different garment styles. Bottom: results generated by TailorNet~\cite{TailorNet_Patel_2020_CVPR}. Note that TailorNet~\cite{TailorNet_Patel_2020_CVPR} needs two templates to perform such garment shapes, and cannot represent some of the shapes. Our method can represent all the shapes in the same framework.}
    \label{fig:exp_cmp_tailor}
\end{figure}

Here, we provide a qualitative comparison with TailorNet~\cite{TailorNet_Patel_2020_CVPR}. Benefiting from our UV-position with mask representation, we can represent different garment shape styles and topologies in the same framework, while TailorNet~\cite{TailorNet_Patel_2020_CVPR} needs separate templates for each kind of garment. As shown in Fig.~\ref{fig:exp_cmp_tailor}, TailorNet~\cite{TailorNet_Patel_2020_CVPR} needs two separate templates for representing T-shirts and shirts; thus, it cannot represent tops, front-opening shirts or half-long-sleeve shirts. Meanwhile, our model can represent all these upper garment shapes in the same model and can perform shape parametrization and reasonable transitions between these shapes, as shown in Fig.~\ref{fig:exp_shape} and our video demo. In addition, TailorNet~\cite{TailorNet_Patel_2020_CVPR} cannot perform shape control for pants to shorts or deal with long dresses, while our method can address such cases.

\textbf{Garment shape inference and editing.} To evaluate our garment shape inference and editing module, we use the 3D scan of the Twindom dataset to perform garment shape inference and editing. The Twindom dataset is a high-resolution 3D scan dataset with multiple clothed humans under arbitrary poses. 

To fit in our module, for the 3D scan clothed model, we first perform a pose alignment procedure with the standard human model to obtain the pose information and the inside posed human mesh $\mathcal{V}_h$, and segment the 3D scan to each garment mesh $\mathcal{V}_g$. We then perform \InferNet introduced in Sect.~\ref{subsec:garment_infer} to extract the shape parameters $s_I$ for the garment, and generate posed sequences using \AnimNet using garment 2D representation generated by $s_I$. The results are shown in Fig.~\ref{fig:teaser}, which shows that we can correctly recover the shape of the original garments. In addition, we can perform shape editing by shifting the 
shape parameters $s_I$ as $s_I'$ and perform the animation procedure. The results are shown in Fig.~\ref{fig:teaser} and Fig.~\ref{fig:exp_infer}.


\textbf{Using SMPL-UV in dress shape representation and animation.}
In our pipeline, we currently use a different UV layout for dress representation and animation. Actually, it is also feasible to represent dresses with SMPL-UV, but there will be artifacts especially on the between-leg regions, when performing dress animation and transition. Here, we perform ablation studies on using SMPL-UV in dress shape representation and animation.

\begin{figure}[ht]
    \centering
    \includegraphics[width=0.85\linewidth]{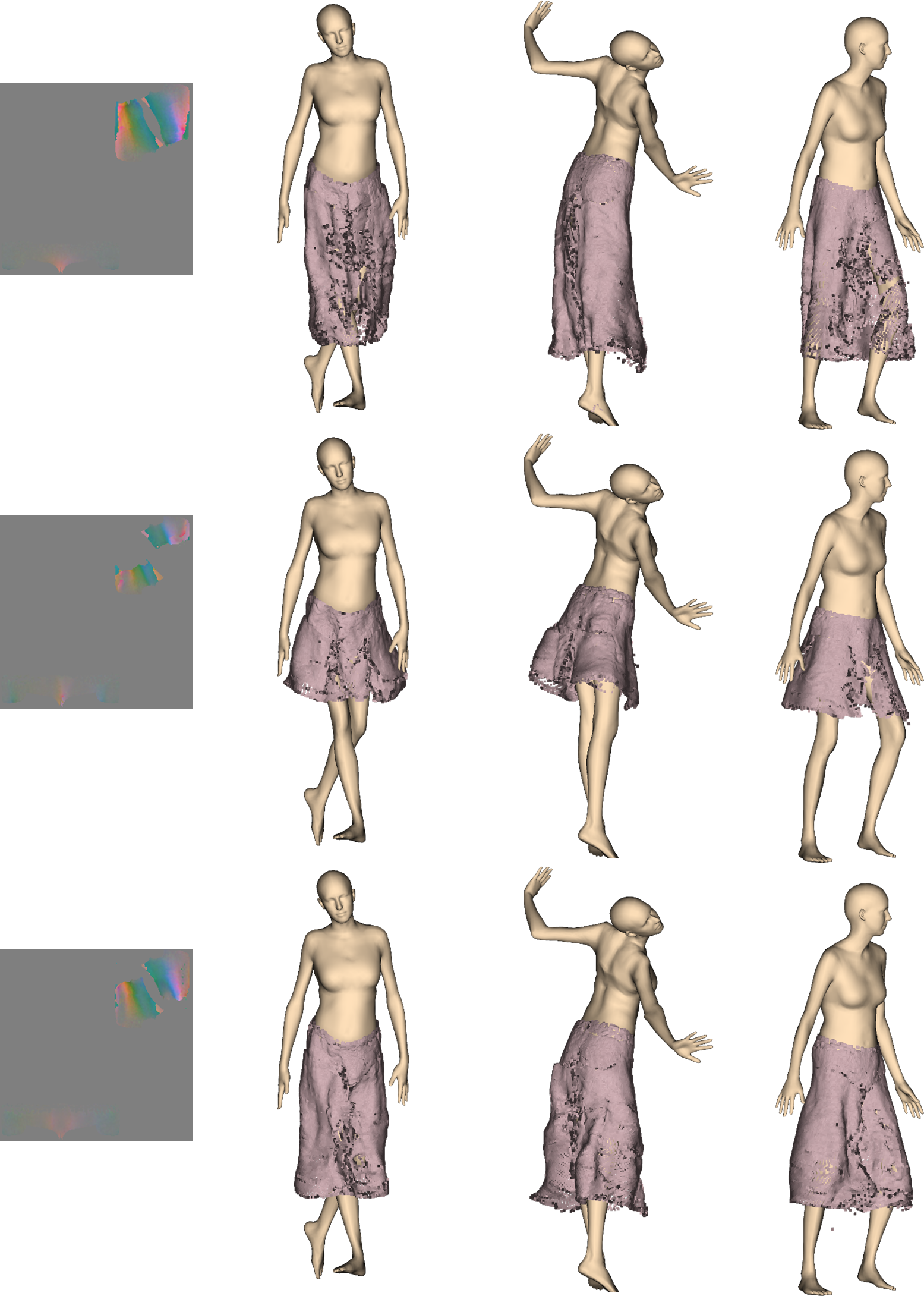}
    \caption{The demonstration of dress representation and animation using SMPL-UV. From left to right: masked dress representation UV map and three animated poses of the corresponding dress.}
    \label{fig:major1_dress}
\end{figure}

To represent a T-posed garment that is not homotopy to the human body, the main concern is to determine the correspondence between the garment and the body model. We solve the SMPL-based surface deformation with the SMPL-garment correspondence by minimizing the Chamfer distance energy function between the corresponding human leg areas on a naked human model and the dress mesh. The UV-position map with continuous $\mathcal{DT}$-based masks is then generated accordingly, similar to Sect.~\ref{subsec:garment_repre}. However, as the dress geometry is not homotopy to the human body, it is difficult to describe as a normal distance map, so we adjust it as a 3-channel shift map, as shown in the left column of Fig.~\ref{fig:major1_dress}. Then, we generate the animated dress geometry UV map similar to Sect.~\ref{subsec:garment_anim} and train our \ParamNet and \AnimNet accordingly. 

As the dresses are not homotopy to the human body, it is difficult to design a mesh layout for completed mesh rendering, so we render the results in a point-based manner. The results are shown in Fig.~\ref{fig:major1_dress} and our video demo, which show that although dress animation under various styled can also be performed using SMPL-UV, there are still artifacts on the between-leg regions and the boundaries. Meanwhile, the non-homotopy dress design will avoid such problems, which is attributed to a more topology-consistent design for the corresponding garment type.

We also perform a quantitative comparison of animation accuracy between the proposed dress-UV and SMPL-UV, as shown in Table~\ref{tab:ablation}. As the SMPL-UV representation better reflects the geometric connection between the SMPL bodies and the dresses, the SMPL-UV performs slightly better quantitatively than the dress-UV. However, the problems addressed above are still difficult to solve. A more general and subtle design may be needed in the future works for representing different garments in the same SMPL-UV.

\begin{figure}[ht]
    \centering
    \includegraphics[width=0.9\linewidth]{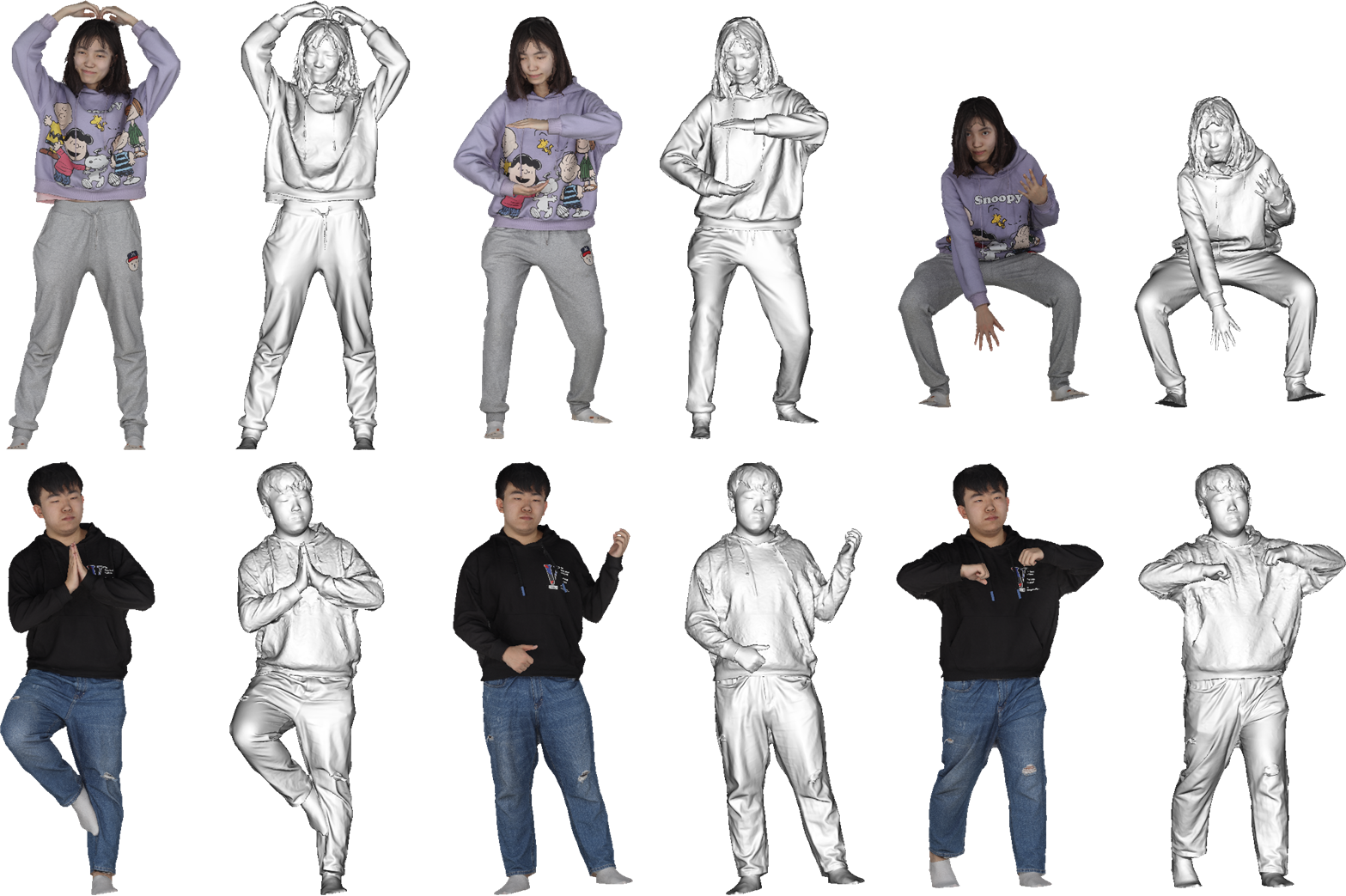}
    \caption{Data samples of the THuman3.0 dataset.}
    \label{fig:exp_major_dataset}
\end{figure}

\begin{figure}[ht]
    \centering
    \includegraphics[width=0.9\linewidth]{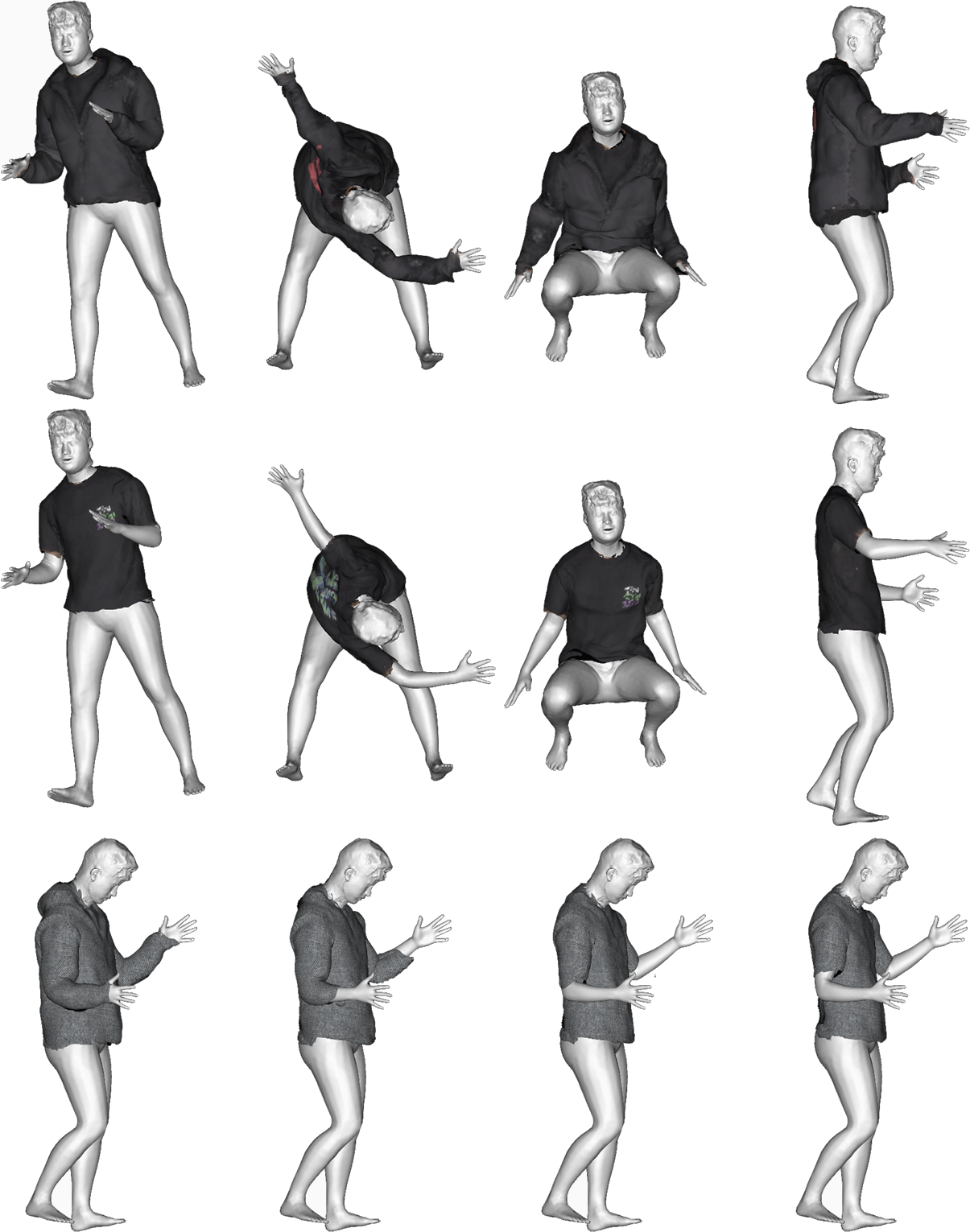}
    \caption{The demonstration of our DeepCloth models tested on THuman3.0 dataset. From top to bottom: animation results of two kinds of garments (long coat / short T-shirt), and garment shape transition between the two garments. Note that our model can perform vivid animation results given different garment styles.}
    \label{fig:exp_major_thuman3.0}
\end{figure}

\textbf{Experiments on real-world data.} 
To illustrate the ability of our model for flexibly representing and animating real-world garments, we also evaluate our model on the real-captured THuman3.0 dataset, which contains 154 human-garment combinations, where each person under 2-3 sets of garments performs 30 to 60 poses. The data samples are shown in Fig.~\ref{fig:exp_major_dataset}. In particular, we evaluate upper garment animation and transition with the given real-captured data. As shown in Fig.~\ref{fig:exp_major_thuman3.0} and our video demo, by performing fine-tuning with the pre-trained \AnimNet and \ParamNet on such a dataset, our model can reflect the dynamic 3D patterns of different garments, and perform vivid animation styles given different upper garments. Additionally, our model can perform smooth and natural style transitions between different garments. The experiments show the ability of our model to deal with real-world garment data. 

\begin{figure}[ht]
    \centering
    \includegraphics[width=0.97\linewidth]{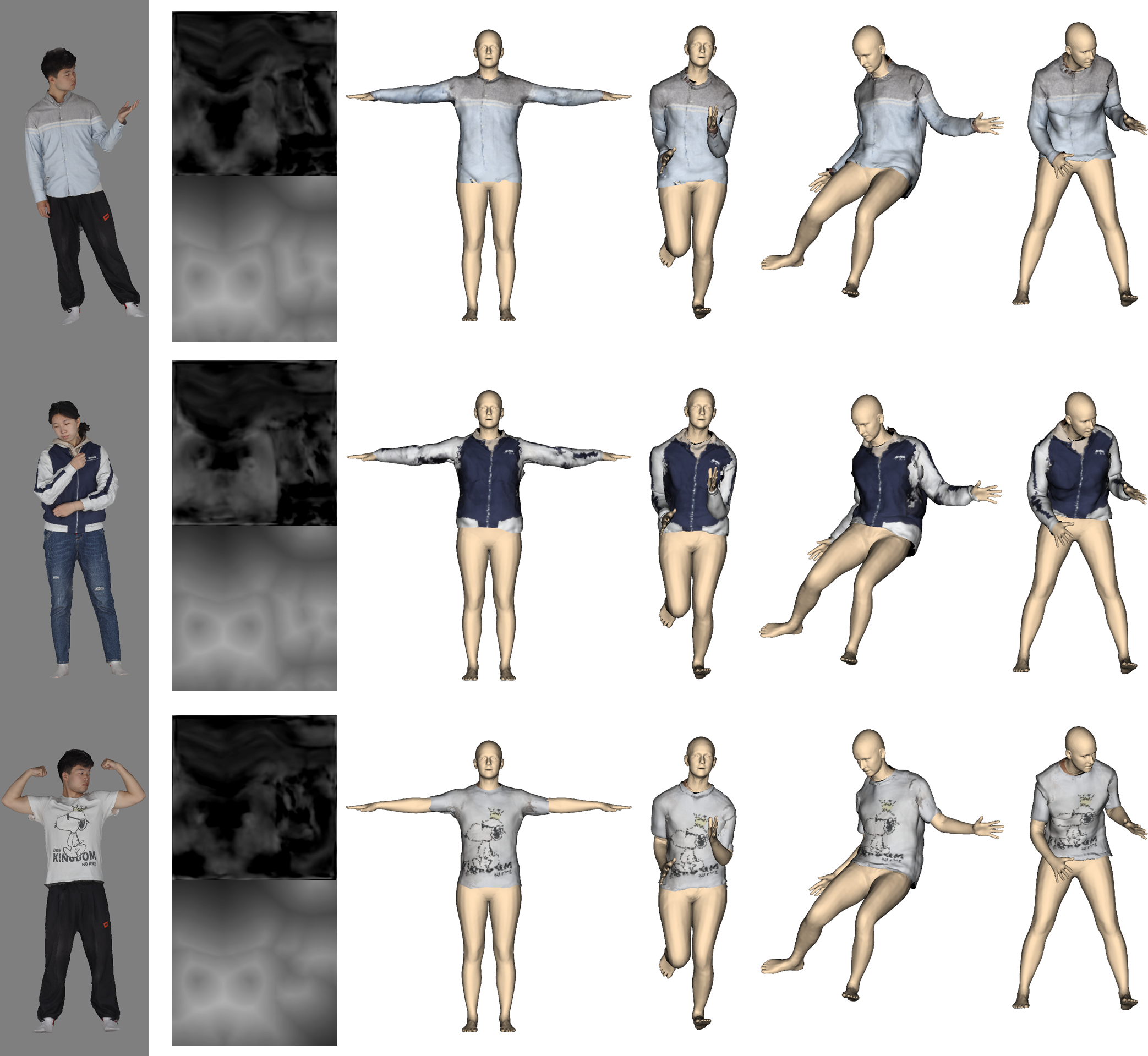}
    \caption{The demonstration of RGB garment reconstruction. From left to right: input real-world RGB image, predicted UV map and $\mathcal{DT}$-mask, reconstructed garment shape, and three animation results.}
    \label{fig:exp_major_rendernet}
\end{figure}

Meanwhile, we also experiment on garment reconstruction from an input RGB image. The network structure is a standard CNN-based encoder-decoder structure, similar to \ParamNet in Fig.~\ref{fig:paramnet}, while the input is replaced by an RGB image. The experiments are also performed on the THuman3.0 dataset. After training, our model can predict the basic garment shapes from an input RGB image in the test set, and the reconstructed garment can be applied to realistic garment animation given garment shape styles, as shown in Fig.~\ref{fig:exp_major_rendernet}. 

\textbf{Quantitative evaluation.} For the garment animation module, we compare our UV-based garment animation method with the PointNet-base~\cite{PointNet_Qi_2017_CVPR} method, which extracts the point features of the garment mesh and the posed human mesh, to infer the shift of garment vertices. As the CLOTH3D~\cite{CLOTH3D_2020_ECCV} dataset contains various garment styles, e.g., front-opening and front-closing T-shirts with long or short sleeves, the garment styles cannot be fit into a fix garment template. Therefore, traditional MLP or other methods suitable for dealing with meshes with a fixed number of vertices could not be evaluated. The CLOTH3D~\cite{CLOTH3D_2020_ECCV} is split into 95 percent for training and 5 percent for testing. The results applied on the test set are as follows; here, the loss is the mean vertex-to-mesh error.

\begin{table}[ht]
    \centering
    \begin{tabular}{c|cc}
    \hline
    garment type           & Ours     & PointNet-based method \\
    \hline
    T-shirts \& shirts     & 16.34    & 20.45                 \\
    pants \& shorts        & 13.51    & 18.63                 \\
    long dresses \& skirts & 31.32    & 40.98                 \\
    long dresses \& skirts (SMPL-UV)  & 27.90 & /             \\
    \hline
    \end{tabular}
    \caption{Mean vertex-to-vertex error (mm) of our \AnimNet method and PointNet-based method for different garment types, with quantitative evaluations on SMPL-UV dresses.}
    \label{tab:ablation}
\end{table}

As shown in Table~\ref{tab:ablation}, our method outperforms the PointNet-based method. This is because the garment styles and topologies vary over a wide range in the CLOTH3D~\cite{CLOTH3D_2020_ECCV} dataset, while traditional PointNet-based~\cite{PointNet_Qi_2017_CVPR} methods have some limits, especially in long dress cases. Note that as our goal is to establish a general garment representation enabling garment shape and style transition, the PointNet-based method actually does not meet our requirement, while our UV-based representation can handle these problems, as demonstrated in Fig.~\ref{fig:exp_param}. 

\begin{figure}[ht]
    \centering
    \subfigure[]{
    \begin{minipage}[t]{0.45\linewidth}
    \centering
    \includegraphics[width=\linewidth]{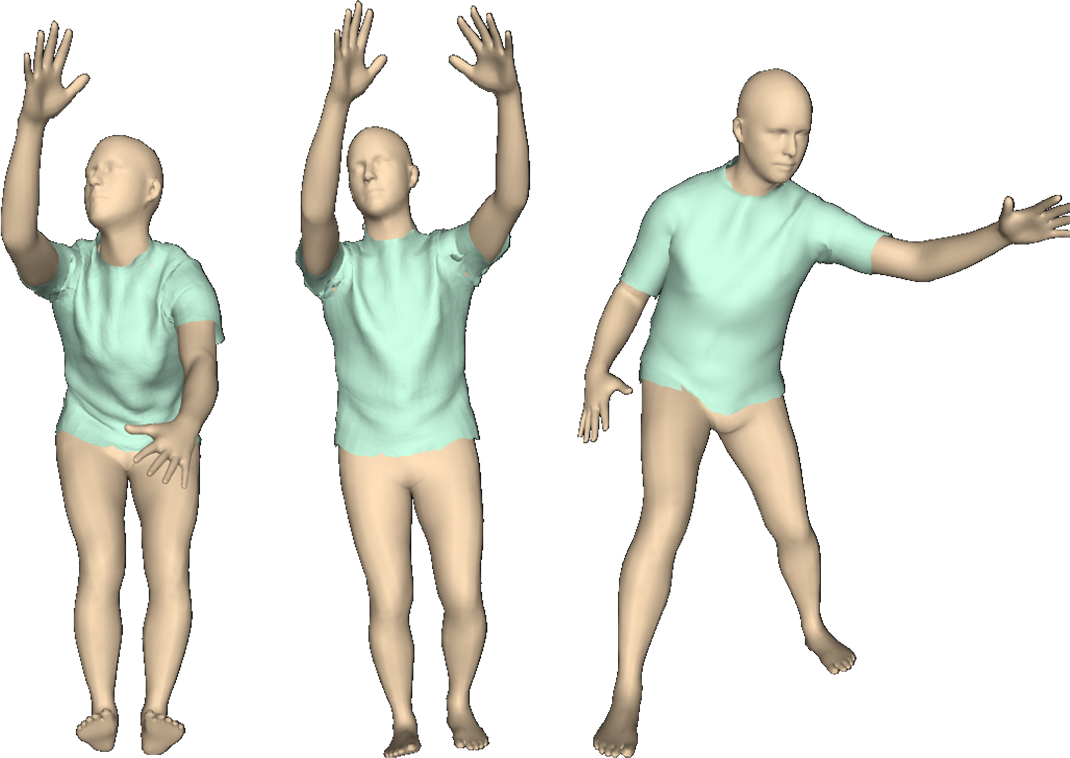}
    \end{minipage}
    }
    \subfigure[]{
    \begin{minipage}[t]{0.45\linewidth}
    \centering
    \includegraphics[width=\linewidth]{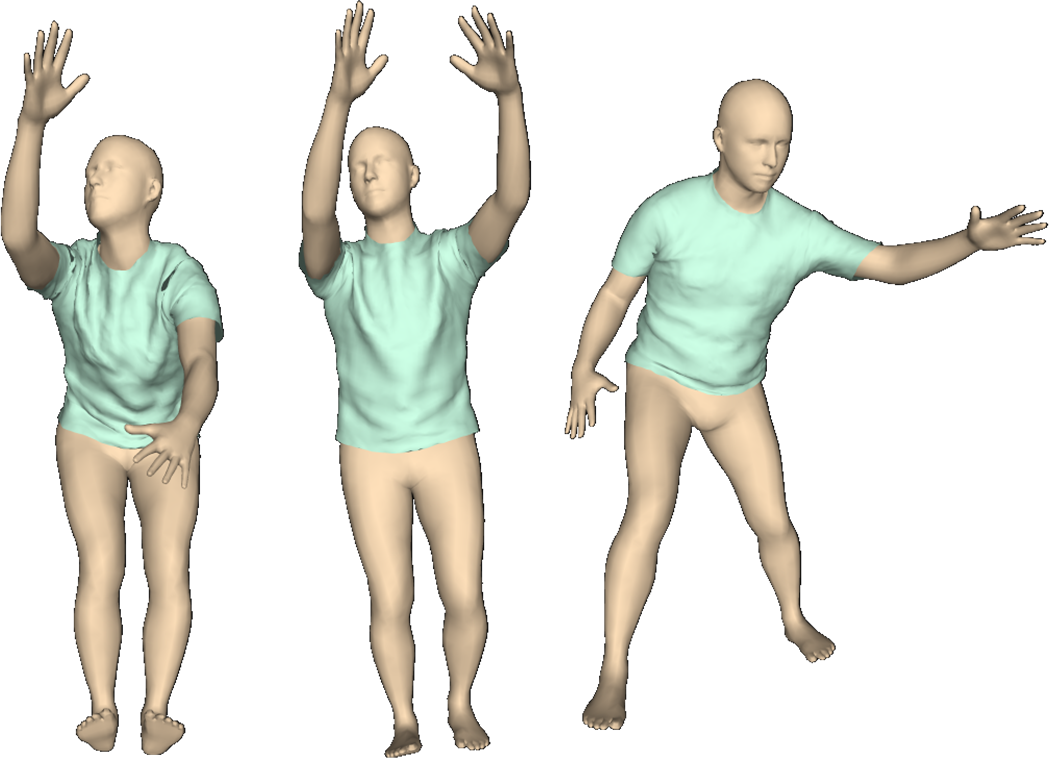}
    \end{minipage}
    }
    \caption{The demonstration of garment animation evaluated on the TailorNet~\cite{TailorNet_Patel_2020_CVPR} dataset: (a) our results, (b) results generated by TailorNet~\cite{TailorNet_Patel_2020_CVPR}.}
    \label{fig:eval_anim_tailor}
\end{figure}

We also evaluate our garment animation module using the TailorNet~\cite{TailorNet_Patel_2020_CVPR} dataset. The dataset we use, i.e., CLOTH3D, does not contain many garment wrinkle details, although it 
contains numerous human pose sequences with different human shapes, each sequence corresponding to an independent garment mesh,
providing multiple garment styles and topologies on both T-pose and animated poses suitable for our framework. In contrast, the TailorNet~\cite{TailorNet_Patel_2020_CVPR} dataset contains garments with more garment wrinkle details, but due to its fixed garment templates for each type of garment, it cannot be used for training garment representation enabling garment topology and style transition. Thus, we only evaluate our garment animation module here. We train our \AnimNet on the TailorNet~\cite{TailorNet_Patel_2020_CVPR} dataset and compare our performance with TailorNet~\cite{TailorNet_Patel_2020_CVPR}. As shown in Fig.~\ref{fig:eval_anim_tailor}, with our framework trained on the TailorNet~\cite{TailorNet_Patel_2020_CVPR} dataset, we can also generate vivid garment details when performing garment animation. Additionally, we make a quantitative evaluation on the TailorNet dataset, as shown in Table~\ref{tab:tailornet}, which shows that by training on a particular garment style, our model can achieve comparable animation qualities with TailorNet. The reason that our model does not achieve better results is that, we focus more on a flexible and general garment representation, while TailorNet focuses more on animation quality and accuracy given fixed garment vertex templates. 

\begin{table}[ht]
    \centering
    \begin{tabular}{c|cc}
    \hline
    garment type           & Ours     & TailorNet method \\
    \hline
    male T-shirts               & 11.58    & 11.2                \\
    male pants                  & 10.39    & 8.1                 \\
    female T-shirts             & 12.97    & 12.3                \\
    female pants                & 6.10     & 4.8                 \\
    \hline
    \end{tabular}
    \caption{Mean vertex-to-vertex error (mm) of our \AnimNet method and TailorNet~\cite{TailorNet_Patel_2020_CVPR} method on the TailorNet dataset for different garment types.}
    \label{tab:tailornet}
\end{table}

\begin{figure}[ht]
    \centering
    \includegraphics[width=0.9\linewidth]{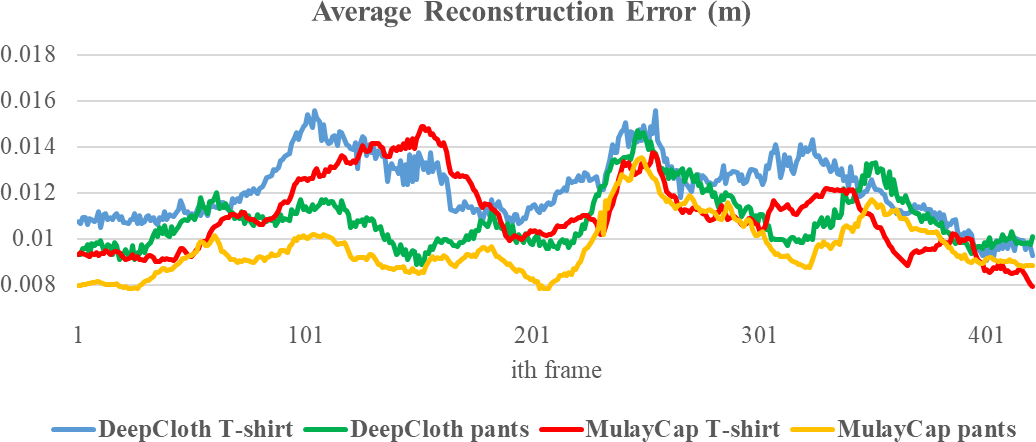}
    \caption{Quantitative comparison with MulayCap~\cite{MulayCap_Su_2020_TVCG} using the rendered 4D model and aligned SMPL poses and shape in the BUFF~\cite{BUFF_Zhang_2017_CVPR} Dataset as input. The result shows a quantitative comparison between the two methods in one 4D sequence using the per-vertex average error.}
    \label{fig:exp_cmp_buff}
\end{figure}

For garment shape inference and application, we compare our method with the state-of-the-art garment reconstruction method MulayCap~\cite{MulayCap_Su_2020_TVCG}, which takes a single-view RGB video as input and dynamically generates a two-layer human with garment mesh. We use a 4D sequence in the BUFF~\cite{BUFF_Zhang_2017_CVPR} dataset as the input, as demonstrated in Fig.~\ref{fig:teaser}. We provide ~\cite{MulayCap_Su_2020_TVCG} with the aligned SMPL shape and poses for every frame, and compare the vertex-to-mesh error between the generated garments and the ground truth input. For our method, we use only the ground truth garment mesh of the first frame for garment shape inference, similar to Fig.~\ref{fig:teaser}, and provide garment animation with SMPL poses and shape. As demonstrated in Fig.~\ref{fig:exp_cmp_buff}, our method performs similarly to MulayCap~\cite{MulayCap_Su_2020_TVCG}. 
From a methodological perspective, MulayCap~\cite{MulayCap_Su_2020_TVCG} is a 4D garment reconstruction pipeline that uses RGB and human parsing information in every frame, for per-frame garment geometry optimization and shape-from-shading geometry detail generation. Our method is an animation module, which only takes the garment mesh of the first frame and the SMPL motion sequence as input, without using the input RGB information, which is why our model can hardly outperform MulayCap~\cite{MulayCap_Su_2020_TVCG}. However, the comparable results still demonstrate the animation ability of our model. 

\section{Conclusion}
\label{sec:conclusion}

In this paper, we propose \ourpaper, a unified neural garment representation framework that can perform garment shape and style transitions by learning the shape space of 3D garments. Our method enables modeling garments under different topologies using the \say{UV-position map with mask} representation, and can perform smooth and free garment transitions by mapping such representations into a continuous feature space. By introducing \AnimNet and \InferNet, our representation allows the generation of 4D clothed human dynamic sequences or the recovery of garment shapes from 3D scans and performing animation and garment shape editing. 
We believe that the proposed topology-aware UV-Mask-based representation takes an important step forward in the field of 3D clothing, especially with the introduction of neural masks for controlling the topology and shape of garments.

\textbf{Limitations and discussions.} Similar to TailorNet~\cite{TailorNet_Patel_2020_CVPR}, we also rely on an explicit collision resolving step. Additionally, at the moment, we cannot handle garments with pockets and collars, which may be resolved by introducing another detailed UV layer. We did not experiment on long dresses that cover the upper body, which can be represented in the future by combining SMPL-UV and dress-UV. Besides above, future works will focus on generating more garment styles based on our work.

\IEEEdisplaynontitleabstractindextext

%
\IEEEpeerreviewmaketitle


%
%
%
%



\ifCLASSOPTIONcaptionsoff
  \newpage
\fi



%

\bibliographystyle{IEEEtran}
\bibliography{bibliography}




%

\begin{IEEEbiography}[{\includegraphics[width=1in,height=1.25in,clip,keepaspectratio]{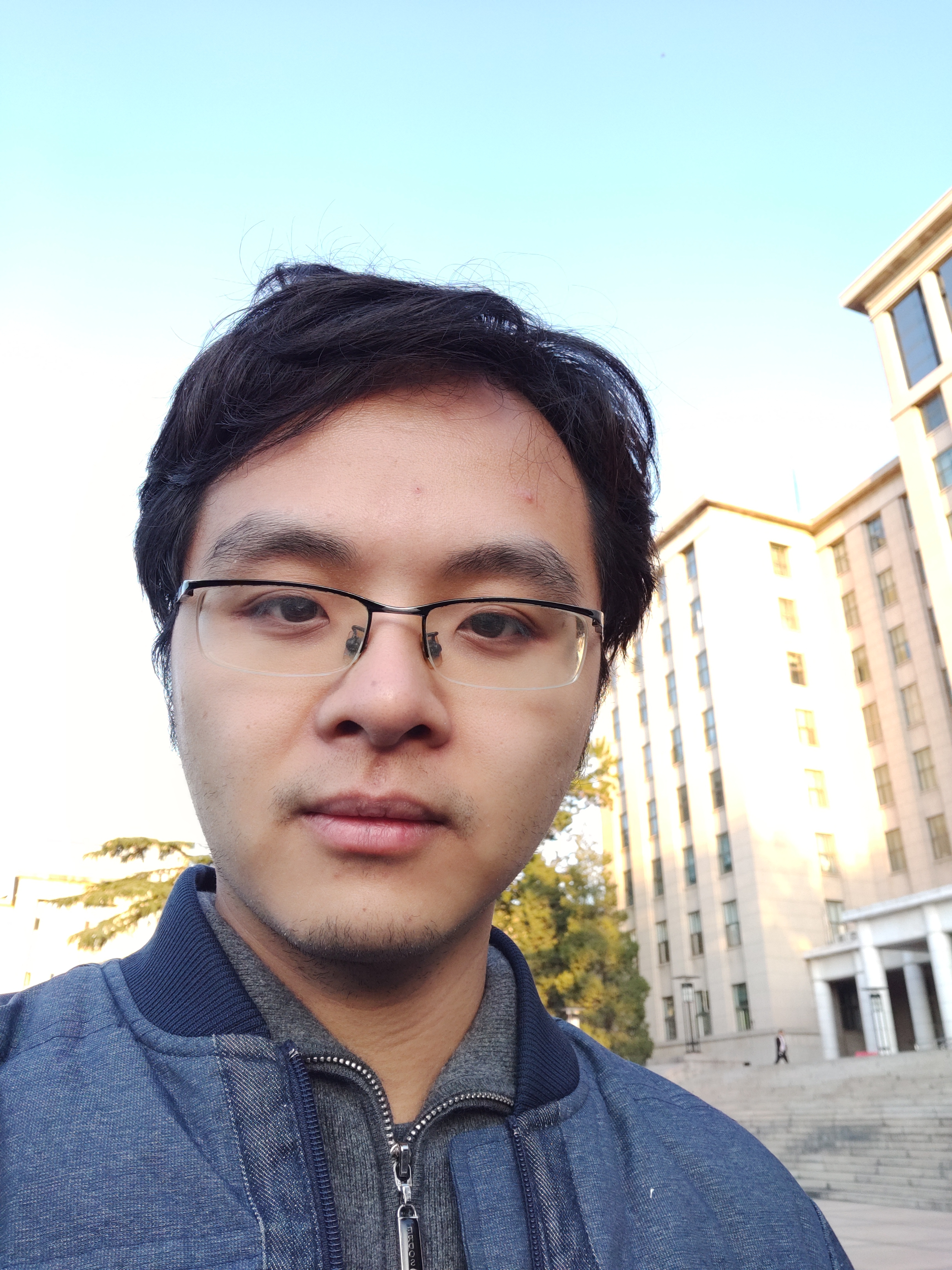}}]{Zhaoqi Su}
received the B.S. degree in Department
of Physics, Tsinghua University, Beijing, China, in 2017.
He is currently pursuing the Ph.D. degree in the
Department of Automation, Tsinghua University, Beijing, China.
\end{IEEEbiography}

\begin{IEEEbiography}[{\includegraphics[width=1in,height=1.25in,clip,keepaspectratio]{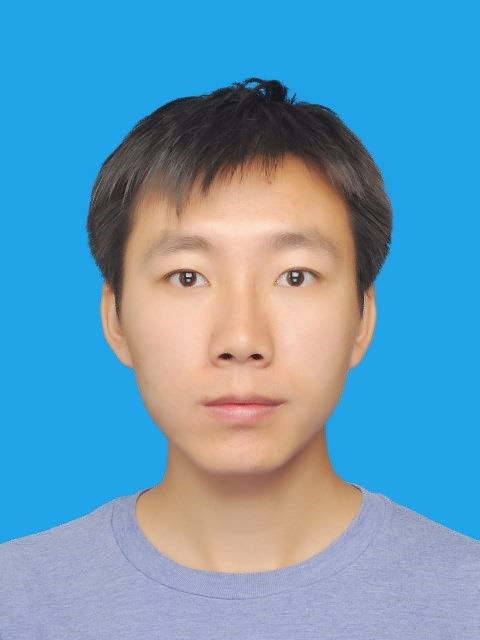}}]{Tao Yu}
is a post-doctoral researcher at Tsinghua University. He received the B.S. degree in Measurement and Control from Hefei University of Technology, China, in 2012, and the Ph.D. degree in instrumental science from Beihang University, China. His current research interests include computer vision and computer graphics.
\end{IEEEbiography}

\begin{IEEEbiography}[{\includegraphics[width=1in,height=1.25in,clip,keepaspectratio]{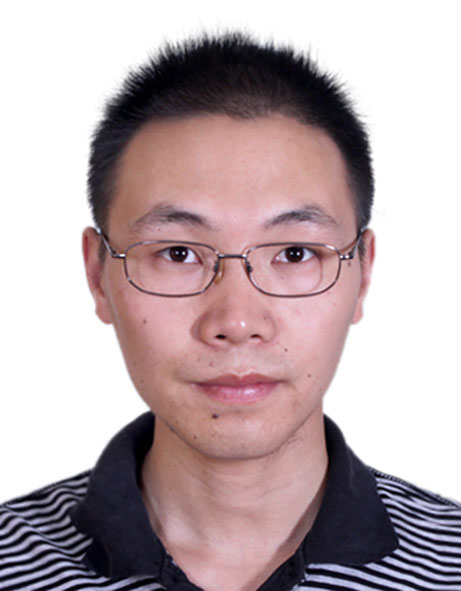}}]{Yangang Wang} received his B.E. degree from Southeast University, Nanjing, China, in 2009 and his Ph.D. degree in control theory and technology from Tsinghua University, Beijing, China, in 2014. He was an associate researcher at Microsoft Research Asia from 2014 to 2017. He is currently an associate professor at Southeast University. His research interests include image processing, computer vision, computer graphics, motion capture and animation.
\end{IEEEbiography}

\begin{IEEEbiography}[{\includegraphics[width=1in,height=1.25in,clip,keepaspectratio]{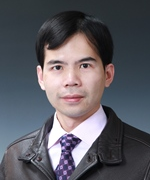}}]{Yebin Liu}
is currently an associate professor at
Tsinghua University. He received the B.E. degree
from the Beijing University of Posts and
Telecommunications, China, in 2002, and the
PhD degree from the Automation Department,
Tsinghua University, Beijing, China, in 2009. He
was a research fellow in the Computer Graphics
Group of the Max Planck Institute for Informatik,
Germany, in 2010. His research areas include
computer vision, computer graphics and computational
photography.
\end{IEEEbiography}




\end{document}